\documentclass[12pt]{article}

\usepackage{arxiv}
\usepackage[utf8]{inputenc}
\usepackage[T1]{fontenc}
\usepackage{url}
\usepackage{subcaption}
\usepackage{xcolor}
\usepackage{booktabs}
\usepackage{amsfonts}
\usepackage{nicefrac}
\usepackage{microtype}
\usepackage{lipsum}
\usepackage{graphicx}
\usepackage{numprint}
\usepackage{titlesec}
\usepackage{XCharter}
\usepackage{amsmath}
\usepackage{adjustbox}
\usepackage{color, colortbl}
\usepackage{amssymb}
\usepackage{bbm}
\usepackage{datetime}
\usepackage[font = {footnotesize, it}]{caption}
\usepackage{tabularx}
\usepackage{rotating}
\usepackage{tablefootnote}
\usepackage{dsfont}
\usepackage{mathtools}
\usepackage{breakcites}
\usepackage{amsthm}
\usepackage{wrapfig}
\usepackage{float}
\usepackage[colorlinks = true, linkcolor = bluecite, urlcolor = bluecite]{hyperref}
\usepackage[outline]{contour}
\usepackage{listofitems}
\usepackage[ruled,vlined]{algorithm2e}
\usepackage{setspace}
\usepackage{siunitx}

\npthousandsep{,}
\npdecimalsign{.}

\DeclareMathOperator*{\argmin}{argmin}

\newcommand{\Esp}[1]{\mathbb{E}\! \left[ #1 \right]}

\definecolor{bluecite}{HTML}{0875b7}
\hypersetup{citecolor = bluecite}

\contourlength{1.4pt}

\colorlet{myred}{red!80!black}
\colorlet{myblue}{blue!80!black}
\colorlet{mygreen}{green!60!black}
\colorlet{myorange}{orange!70!red!60!black}
\colorlet{mydarkred}{red!30!black}
\colorlet{mydarkblue}{blue!40!black}
\colorlet{mydarkgreen}{green!30!black}
\colorlet{mygray}{gray!80}
\colorlet{mygray_pale}{gray!30}

\usepackage{tikz,graphicx,pgfplots,tikz-layers, fontawesome5}
\pgfplotsset{compat=1.18} 
\usetikzlibrary{positioning,matrix,fit,decorations,decorations.text, arrows,arrows.meta, calc}
\usepackage{amssymb,amsmath,enumitem,lipsum}
\usepackage{multirow,fancyhdr}
\usepackage{array,xcolor}

\title{Telematics Combined Actuarial Neural Networks for Cross-Sectional and Longitudinal Claim Count Data}

\author{
    Francis Duval \\
    Chaire Co-operators en analyse des risques actuariels\\
    Département des mathématiques\\
    Université du Québec à Montréal\\
    Montréal, QC H2X 3Y7\\
    \texttt{duval.francis.2@courrier.uqam.ca} \\
    \And
    Jean-Philippe Boucher\\
    Chaire Co-operators en analyse des risques actuariels\\
    Département des mathématiques\\
    Université du Québec à Montréal\\
    Montréal, QC H2X 3Y7\\
    \texttt{boucher.jean-philippe@uqam.ca} \\
    \And
    Mathieu Pigeon\\
    Chaire Co-operators en analyse des risques actuariels\\
    Département des mathématiques\\
    Université du Québec à Montréal\\
    Montréal, QC H2X 3Y7\\
    \texttt{pigeon.mathieu.2@uqam.ca} \\
}

\begin{document}

\setstretch{1.25}

\maketitle

\begin{abstract}
We present novel cross-sectional and longitudinal claim count models for vehicle insurance built upon the Combined Actuarial Neural Network (CANN) framework proposed by Mario Wüthrich and Michael Merz. The CANN approach combines a classical actuarial model, such as a generalized linear model, with a neural network. This blending of models results in a two-component model comprising a classical regression model and a neural network part. The CANN model leverages the strengths of both components, providing a solid foundation and interpretability from the classical model while harnessing the flexibility and capacity to capture intricate relationships and interactions offered by the neural network. In our proposed models, we use well-known log-linear claim count regression models for the classical regression part and a multilayer perceptron (MLP) for the neural network part. The MLP part is used to process telematics car driving data given as a vector characterizing the driving behavior of each insured driver. In addition to the Poisson and negative binomial distributions for cross-sectional data, we propose a procedure for training our CANN model with a multivariate negative binomial (MVNB) specification. By doing so, we introduce a longitudinal model that accounts for the dependence between contracts from the same insured. Our results reveal that the CANN models exhibit superior performance compared to log-linear models that rely on manually engineered telematics features. 
\end{abstract}


\section{Introduction and Motivations}\label{sec:intro}

Vehicle insurance products have traditionally been priced based on self-reported attributes provided by insureds. These attributes commonly include various risk factors, including gender, age, vehicle usage, and claim history. Insurers rely on this information to assess the level of risk associated with each insurance contract and determine appropriate premium rates. With the introduction of telematics technology, insurers can now collect a wide range of driving data through devices installed in the vehicles of policyholders or through mobile applications. This includes information such as vehicle speed, acceleration and braking behavior, mileage, location data, and factors like the time of day or types of roads frequently traveled. By leveraging this wealth of data, insurers can gain a more accurate and objective understanding of each individual's driving habits and style, enabling them to customize insurance offerings and pricing based on their actual driving behavior. This emerging paradigm, known as Usage-Based Insurance (UBI), revolutionizes the insurance landscape in various ways. For insurers, telematics data means more accurate risk assessment algorithms, which can often translate into a competitive advantage. For insureds, it means fairer premium rates that align more closely with their actual risk profiles rather than being computed based on broad demographic categories. It also means that they are priced based on risk indicators over which they have control. From a societal perspective, UBI also offers many advantages. One of the key benefits is the potential to improve road safety. By giving incentives for safe driving behavior and reduced mileage, UBI not only helps reduce the frequency and severity of accidents, ultimately saving lives and reducing the economic burden associated with road accidents, but also contributes to reducing greenhouse gas emissions. Additionally, telematics provide insurers with viable alternatives to sensitive risk factors, thereby helping to prevent unfair discrimination. For a more extensive overview of the benefits of UBI, we refer to the works of \cite{litman2007}, \cite{bordoff2008} and \cite{ziakopoulos2022transformation}.

One of the most prominent questions related to UBI is how to make the most out of the collected driving data. A significant subset of the literature has focused on incorporating mileage into pricing models due to its acknowledged importance as a risk factor in assessing risk and determining premium rates (see, for instance, \cite{boucher2017exposure}, \cite{lemaire2015use}, and \cite{turcotte2023gamlss}). However, mileage alone fails to provide the whole story about an insured individual's driving behavior, prompting researchers to consider additional telematics information. One prevalent approach involves drawing upon domain knowledge to craft telematics features from raw data. By applying their expertise in the field, researchers can engineer features that capture critical aspects of driving behavior, specifically driving characteristics that are thought to be correlated with the risk of accident. Common examples of such features include harsh braking/acceleration events, cornering events, speeding, distracted driving, the fraction of driving during both different time slots (e.g., rush hour, late-night hours, weekdays), and on different road types (e.g., urban roads, highways), as well as the fraction of driving in different speed slots. While this approach captures signals missed by traditional risk factors and mileage (thereby improving pricing accuracy), it relies heavily on human judgment, with its inherent flaws and biases. With countless possible telematics features that can be engineered from raw telematics data, selecting the optimal ones for pricing is not straightforward. Furthermore, this process necessitates the setting of thresholds. For example, how should night driving or harsh braking be precisely defined?

The limitations of the aforementioned approach have motivated researchers to explore a new set of methods that rely more on data and decrease the need for human judgment. As highlighted in a recent study by \cite{embrechts2022recent}, the increasing amount of data available presents a challenge in manually designing features, leading actuaries to increasingly depend on tools like neural networks to learn and extract meaningful representations from the data. \cite{blier2021rethinking} underline the importance of learning valuable representations from emerging data sources such as text, image, and sensor data. These sources, which include telematics car driving data, can enrich traditional data and offer improved insights for predicting future losses in insurance contracts. Neural networks are regarded as the most effective means for automatically extracting valuable features from raw data, which validates their practical application. In recent years, researchers have successfully applied the toolbox of deep learning, namely neural networks architectures with a large number of hidden layers, to handle telematics data and other types of unstructured data. In their work, \cite{wuthrich2017covariate} introduce the speed-acceleration heatmap, a matrix representation that characterizes the driving style of an insured driver, which is well-suited for processing by deep learning algorithms. Subsequent studies (\cite{gao2018feature}, \cite{gao2019claims}, \cite{gao2019convolutional}, \cite{gao2022boosting}) have effectively leveraged these heatmaps by employing neural networks to learn representations from them. In \cite{meng2022improving}, the authors propose a supervised driving risk scoring convolutional neural network model that uses telematics car driving data to improve automobile insurance claims frequency prediction. \cite{blier2020encoding} propose a Convolutional Regional Autoencoder model for generating geographical risk encodings using convolutional neural networks. The resulting encodings, which aim to replace the traditional territory variable, proved beneficial for risk-related regression tasks.

In this paper, we present novel claim count models based on the Combined Actuarial Neural Network (CANN) approach, initially proposed by \cite{wuthrich2019yes}. The CANN approach involves embedding a classical regression model, such as a generalized linear model (GLM; see \cite{nelder1972generalized} and \cite{dionne1989generalization}), into a neural network, achieved by blending the regression functions of both models. Consequently, the resulting model comprises the two following components: the classical regression (or actuarial) model and the neural network. This blending process can be interpreted as a form of neural network boosting for the actuarial model, combining the strengths of both approaches. The calibration of the CANN neural network is performed using the classical actuarial model as the initial value in the gradient descent algorithm, with the negative log-likelihood of the specified distribution used as the loss function. One of the key benefits of this specific architecture is the solid foundation offered by the classical model, complemented by the network component's flexibility and pattern recognition capabilities. Neural networks excel in approximating highly nonlinear functions and possess the ability to compute valuable interactions between input variables automatically. Consequently, the CANN approach combines the best of both worlds, leveraging the interpretability and reliability of the classical model while capitalizing on the power of neural networks to capture complex relationships and patterns in the data. A few studies have successfully leveraged this approach: \cite{schelldorfer2019nesting} present a case study where a Poisson GLM for predicting claims frequencies is initially used, then enhanced through generalized additive models (GAMs) with natural cubic splines and finally combined with a neural network, resulting in a CANN approach. The study also explores the use of embedding layers for more efficient treatment of categorical variables; \cite{gabrielli2020neural} boost an overdispersed Poisson model with a multilayer perceptron to improve individual loss reserving; \cite{tzougas2023enhancing} use the CANN approach to enhance binary classification; \cite{laporta2023neural} apply the CANN architecture in the context of quantile regression.

Our models employ a log-linear model for the actuarial model part and a multilayer perceptron (MLP) for the network part. Telematics information is incorporated into the MLP as a telematics vector, which is given as input to represent the driving behavior of each insured driver. The MLP part additionally includes traditional risk factors as inputs, enabling interactions between traditional and telematics inputs, while the log-linear part, constrained in estimating complex functions, is only given traditional risk factors. We explore three distinct distribution specifications for the claim count: Poisson and negative binomial for cross-sectional analysis, and multivariate negative binomial (MVNB; see \cite{hausman1984} and \cite{boucher2008models}), also known as \textit{negative multinomial}, for longitudinal analysis. The MVNB distribution is a popular choice for modeling longitudinal claim count data, as it captures the dependence between contracts from the same insured. However, to our knowledge, this specification has never been adapted to a neural network model for claim count regression. In this study, we extend the application of the MVNB distribution by incorporating it into the neural network framework, specifically the CANN architecture, for modeling longitudinal claim count data. This adaptation allows us to leverage the strengths of both the MVNB distribution and the neural network architecture. Our findings indicate that the CANN models perform better than their log-linear counterparts that rely on manually engineered telematics features. Furthermore, the CANN model using the MVNB specification exhibits a significant improvement compared to the two cross-sectional specifications. 

In Section~\ref{sec:data}, we present the two datasets available to us: the contract dataset and the telematics dataset. Following that, in Section~\ref{sec:models}, we delve into the theory behind the CANN claim count models and also discuss the log-linear models that serve as benchmarks. Moving on to Section~\ref{sec:application}, we provide an explanation of how we apply the models on our specific dataset and show how we preprocess telematics data. In Section~\ref{sec:analyzes}, we assess the performance of the models on a holdout sample and interpret the CANN models through permutation feature importance and partial dependence plots. Lastly, we conclude in Section~\ref{sec:conclusions_article_3}.

\section{Data}\label{sec:data}

We have access to data from a Canadian property and casualty insurance company, which comes in two distinct datasets: the contract and the telematics dataset. 

\subsection{Contract dataset}

\begin{figure}[ht]
    \centering
    \includegraphics[width = \textwidth]{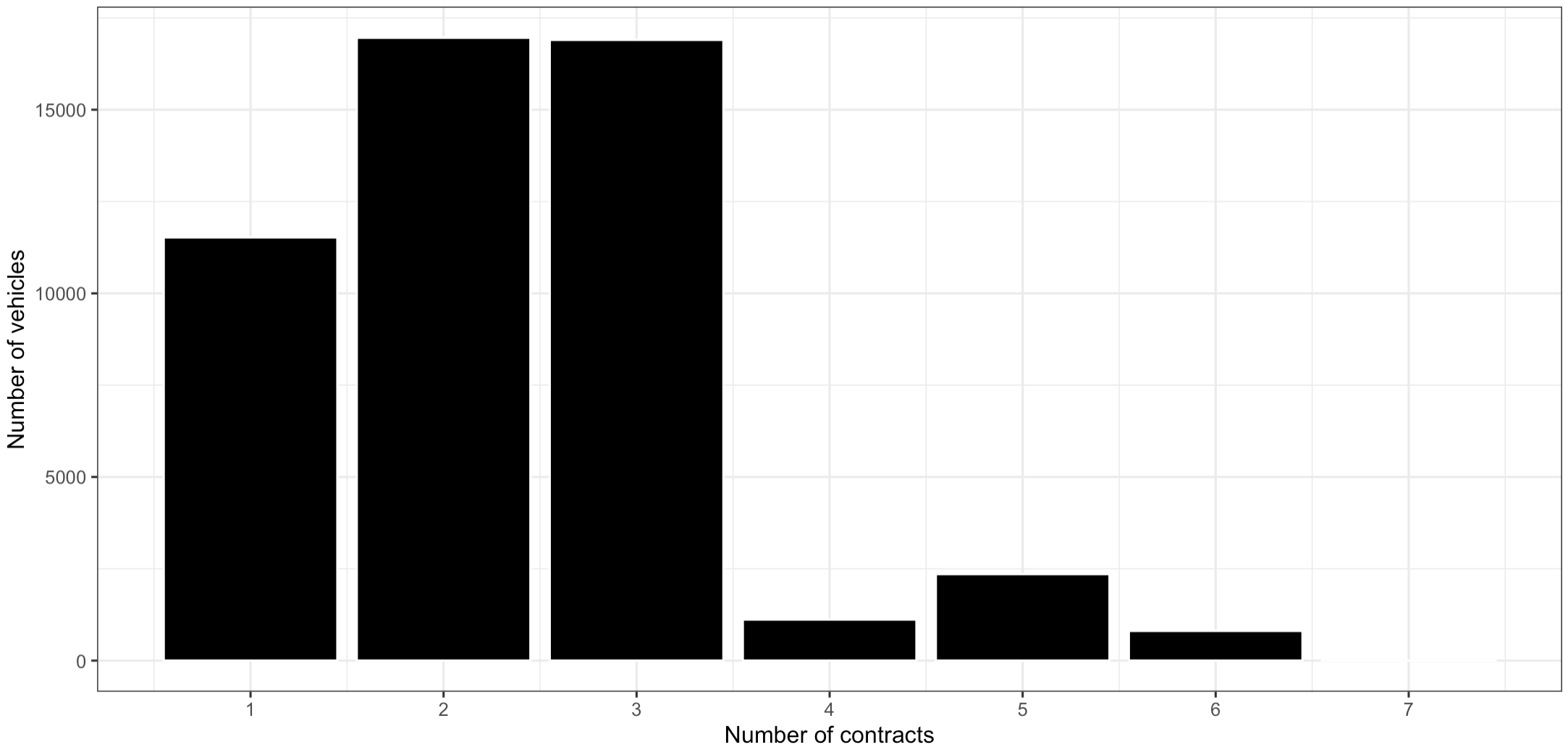}
    \caption{Number of contracts per vehicle.}
    \label{fig:plot_nb_contracts_per_vehicle}
\end{figure}

In the contract dataset, each row represents a unique insurance contract. Contracts typically last for one year, but there are instances where their duration may be shorter or longer. Each vehicle is observed over one or more contracts; therefore, one vehicle can be represented by one or more rows in this dataset. Based on risk factors, a premium must be computed for each contract. When using a cross-sectional data model, contracts from the same vehicle are assumed to be independent of each other. On the other hand, a longitudinal data model assumes a dependence between contracts, allowing it to use information from previous contracts (including traditional risk factors, telematics data, past claims, etc.) to compute the premium. The contract dataset includes attributes commonly used in vehicle insurance pricing models. These traditional risk factors, displayed in Table~\ref{tab:classic}, are recorded for \numprint{117268} insurance contracts initiated between December $\text{15}^\text{th}$, 2015 and December $\text{31}^\text{st}$, 2018. 
\begin{table}[ht]
    \centering
    \begin{adjustbox}{max width = \textwidth}
        \begin{tabular}{l l l}
            \toprule 
            \textbf{Variable name} & \textbf{Description} & \textbf{Type} \\
            \midrule
            \texttt{vin} & Unique vehicle identifier & ID\\
            \cmidrule(l){1-3}
            \texttt{annual\_distance} & Annual distance declared by the insured & Numeric\\
            \texttt{commute\_distance} & Distance to the place of work declared by the insured & Numeric\\
            \texttt{conv\_count\_3\_yrs\_minor} & Number of minor contraventions in the last three years & Numeric\\
            \texttt{distance} & Real distance driven & Numeric\\
            \texttt{expo} & Contract duration in years & Numeric\\
            \texttt{gender} & Gender of the insured & Categorical\\
            \texttt{marital\_status} & Marital status of the insured & Categorical\\
            \texttt{pmt\_plan} & Payment plan chosen by the insured & Categorical\\
            \texttt{veh\_age} & Vehicle age & Numeric\\
            \texttt{veh\_use} & Use of the vehicle & Categorical\\
            \texttt{years\_licensed} & Number of years since obtaining driver's license & Numeric\\
            \cmidrule(l){1-3}
            \texttt{nb\_claims} & Number of claims & Numeric\\
            \bottomrule 
        \end{tabular}
    \end{adjustbox}
    \caption{Variables of the contract dataset.} 
    \label{tab:classic} 
\end{table}
In cases where multiple drivers are associated with a particular contract, attributes of the principal driver are used. Additionally, the dataset includes the vehicle identification number (VIN), allowing us to identify the insured vehicle accurately, alongside the reported claim count. As our goal is to perform claim count regression on contracts, the claim count variable will serve as the response for our supervised learning algorithms, namely the log-linear and the CANN models. The \numprint{117268} contracts are associated with \numprint{49671} distinct vehicles, resulting in an average of approximately \numprint{2.36} contracts per vehicle. The histogram of the number of contracts per vehicle is shown in Figure~\ref{fig:plot_nb_contracts_per_vehicle}.

\subsection{Telematics dataset}

All \numprint{117268} contracts have been logged using an on-board diagnostics (OBD) device, capturing driving information. This data is stored as trip summaries in the telematics dataset, which comprises \numprint{117566259} trips. Each row in the dataset represents a specific trip, and every trip is described by \numprint{4} attributes: the departure and arrival date and time, the distance driven, and the maximum speed reached. Additionally, each trip is associated with a VIN, and with the date information, it is thus possible to link each trip with one of the \numprint{117268} contracts. An extract from the telematics dataset is presented in Table~\ref{tab:telematics}.
\begin{table}[ht]
    \centering
    \begin{adjustbox}{max width = \textwidth}
        \begin{tabular}{c c c c c c}
            \toprule 
            \textbf{VIN} & \textbf{Trip ID} & \textbf{Departure datetime} & \textbf{Arrival datetime} & \textbf{Distance} & \textbf{Maximum speed}\\ 
            \midrule
            A & $1$ & $2017$-$05$-$02$ {$19$:$04$:$15$} & $2017$-$05$-$02$ {$19$:$24$:$24$} & $25.0$ & $104$\\
            A & $2$ & $2017$-$05$-$02$ {$21$:$31$:$29$} & $2017$-$05$-$02$ {$21$:$31$:$29$} & $6.4$ & $66$\\
            \vdots & \vdots & \vdots & \vdots & \vdots & \vdots \\
            A & $2320$ & $2018$-$04$-$30$ {$21$:$17$:$22$} & $2018$-$04$-$30$ {$21$:$18$:$44$} & $0.2$ & $27$\\
            \cmidrule(l){1-6}
            B & $1$ & $2017$-$03$-$26$ {$11$:$46$:$07$} & $2017$-$03$-$26$ {$11$:$53$:$29$} & $1.5$ & $76$\\
            B & $2$ & $2017$-$03$-$26$ {$15$:$18$:$23$} & $2017$-$03$-$26$ {$15$:$51$:$46$} & $35.1$ & $119$\\
            \vdots & \vdots & \vdots & \vdots & \vdots & \vdots \\
            B & $1485$ & $2018$-$03$-$23$ {$20$:$07$:$08$} & $2018$-$03$-$23$ {$20$:$20$:$30$} & $10.1$ & $92$\\
            \cmidrule(l){1-6}
            C & $1$ & $2017$-$11$-$20$ {$08$:$14$:$34$} & $2017$-$11$-$20$ {$08$:$40$:$21$} & $9.7$ & $78$ \\
            \vdots & \vdots & \vdots & \vdots & \vdots & \vdots \\
            \bottomrule 
        \end{tabular}
    \end{adjustbox}
    \caption{Extract from the telematics dataset. Dates are displayed in the yyyy-mm-dd format. The actual VINs have been hidden for privacy purposes.} 
    \label{tab:telematics}
\end{table}

\subsection{Training, validation, and testing datasets}

In supervised learning analysis, splitting the available data into training, validation, and testing sets is paramount for ensuring the reliability and ability to generalize of the learned model. The training set, which usually comprises the largest portion of the data, is used to train the model's parameters and optimize its performance. However, relying solely on the training set for performance assessment can lead to overfitting, particularly when the model has a high capacity. To address this, the validation set is used during the modeling process to assess the model's performance on unseen data. It plays an important role in tuning hyperparameters, selecting the optimal model architecture, and preventing overfitting. By assessing the model's performance on the validation set, one can obtain an estimate of its generalization performance and make necessary adjustments to improve its ability to generalize well to new, unseen examples. However, it is important to note that the back-and-forth process of evaluating the model on the validation set and adjusting its hyperparameters can introduce information leakage from the validation set into the training set. This can create an illusion of better performance than the model would exhibit in real-world scenarios. As a result, the testing set is reserved for the final evaluation of the learned model. It serves as an unbiased assessment of how well the model will perform on completely unseen data. This final evaluation provides an estimate of the model's true performance and helps determine its reliability in real-world scenarios. By keeping the testing set separate from the training and validation sets, we can ensure an unbiased evaluation and avoid any potential data leakage. We partition the data as outlined in Table~\ref{tab:split} for our analysis. Approximately \numprint{60}\% of the vehicles are allocated for training, while approximately \numprint{20}\% is assigned to the validation and testing sets.
\begin{table}[ht]
    \centering
    \begin{adjustbox}{max width = \textwidth}
        \begin{tabular}{l c c c c}
            \toprule 
            \textbf{Set} & \textbf{Symbol} & \textbf{Number of vehicles} & \textbf{Number of contracts} & \textbf{Number of trips}\\ 
            \midrule
            Training & $\mathcal{T}_r$ & $\numprint{30000}$ & $\numprint{70451}$ & $\numprint{71416560}$\\
            Validation & $\mathcal{V}_a$ & $\numprint{10000}$ & $\numprint{23368}$ & $\numprint{22611829}$\\
            Testing & $\mathcal{T}_e$ & $\numprint{9671}$ & $\numprint{23449}$ & $\numprint{23537870}$\\
            \midrule
            Total & -- & $\numprint{49671}$ & $\numprint{117268}$ & $\numprint{117566259}$\\
            \bottomrule 
        \end{tabular}
    \end{adjustbox}
    \caption{Data partitioning} 
    \label{tab:split}
\end{table}

\section{Count Regression Models}\label{sec:models}

We consider a training dataset denoted as $\mathcal{T}_r$, which consists of $|\mathcal{T}_r|$ rows representing vehicle insurance contracts. Contracts are grouped by vehicle and each vehicle $i$ is observed over $T_i$ contracts. We define $Y_{it}$ as a discrete random variable denoting the number of claims during the $t^\text{th}$ contract of vehicle $i$. Furthermore, we have $\boldsymbol{x}_{it}$ a vector containing relevant predictor variables associated with the $t^\text{th}$ contract of vehicle $i$. Importantly, we assume independence among all insured vehicles. In claim count regression, the ultimate goal is to estimate the probability mass function (PMF) of the number of claims, given all past and current information about the vehicle. Mathematically, we seek to estimate:
\begin{align}
    \mathbb{P}\left(Y_{it} = y_{it}| \boldsymbol{y}_{i, (1:t-1)}, \boldsymbol{x}_{i, (1:t)}\right), \quad y_{it} \in \mathbb{N},
\end{align}
where $\boldsymbol{y}_{i, (1:t-1)} = (y_{i1}, \dots, y_{i, t-1})$ is the vector of past claims and $\boldsymbol{x}_{i, (1:t)} = \{\boldsymbol{x}_{i1}, \dots, \boldsymbol{x}_{it}\}$ is the set of past and current covariate vectors for vehicle $i$.

\subsection{Cross-sectional models}

In addition to assuming independence between vehicles, cross-sectional models also assume independence between contracts from the same vehicle. Consequently, these models do not use the history of a vehicle to estimate its future risk. The PMF of the number of claims can thus be written as:
\begin{align}
    \mathbb{P}\left(Y_{it} = y_{it}| \boldsymbol{y}_{i, (1:t-1)}, \boldsymbol{x}_{i, (1:t)}\right) = \mathbb{P}\left(Y_{it} = y_{it}| \boldsymbol{x}_{it}\right), \quad y_{it} \in \mathbb{N}.
\end{align}

\subsubsection{Poisson regression}

The Poisson distribution is widely used in supervised learning analysis for claim count data due to its good properties and simplicity. Under the Poisson specification, the PMF of the claim count for the $t^\text{th}$ contract of vehicle $i$, denoted by $Y_{it}$, given its predictor vector, denoted by $\boldsymbol{x}_{it}$, is defined by
\begin{align}
    \mathbb{P}(Y_{it} = y_{it}|\boldsymbol{x}_{it}) = \frac{e^{-\mu(\boldsymbol{x}_{it})}\mu(\boldsymbol{x}_{it})^{y_{it}}}{y_{it}!},\quad \text{for} \quad y_{it} \in \mathbb{N},
    \label{eq:poisson}
\end{align}
with $\Esp{Y_{it}|\boldsymbol{X}_{it} = \boldsymbol{x}_{it}} = \text{Var}[{Y_{it}|\boldsymbol{X}_{it} = \boldsymbol{x}_{it}}] = \mu(\boldsymbol{x}_{it})$. The mean parameter $\mu(\boldsymbol{x}_{it})$ denotes the conditional expectation (and conditional variance) of $Y_{it}$. The regression function $\mu(\cdot)$ captures the relationship between the predictors $\boldsymbol{x}_{it}$ and the mean parameter in the Poisson distribution, indicating how the conditional expected count is influenced by the predictors. Subsequently, one must choose a specific functional form for $\mu(\cdot)$, which defines a hypothesis function space $\mathcal{H}$ that includes all the candidate functions for modeling $\mu(\cdot)$. The next step involves selecting the optimal function $\widehat{\mu} \in \mathcal{H}$, equivalent to estimating the parameters of the specified functional form based on the available data.

In order to define what constitutes a good regression function, it is necessary to select a suitable loss function that quantifies the dissimilarity between the estimated probability mass and the true label. The goal is to minimize this dissimilarity, improving the model’s predictive performance. The cross-entropy loss, also known as the negative log-likelihood loss, is a commonly chosen option. For a specific observation $i$, the cross-entropy loss is given by $-\ln(p_i)$, where $p_i$ is the estimated probability of observing the true label $y_i$. This loss function assigns a higher penalty to larger discrepancies between the true label and the predicted probability, incentivizing the model to converge towards more accurate predictions. To estimate the parameters, we typically aim to minimize the average loss function over the training set, also called the \textit{empirical risk}. In the case of Poisson regression, this involves minimizing the average Poisson cross-entropy by solving the following optimization problem:
\begin{align}
    \widehat{\mu} = \argmin_{\mu \in \mathcal{H}} \left\{ -\frac{1}{|\mathcal{T}_r|} \sum_{(i, t)\in\mathcal{T}_r} y_{it} \ln[\mu(\boldsymbol{x}_{it})] - \mu(\boldsymbol{x}_{it}) - y_{it}! \right \}.
    \label{eq:loss_poisson}
\end{align}
Note that this is equivalent to maximizing the likelihood function. For some specifications of $\mu(\cdot)$, notably the log-linear specification, the criterion in Equation~(\ref{eq:loss_poisson}) is convex, which enables various convex optimization techniques to be applied. Alternative estimation techniques can also be used. One common option is regularization techniques, including lasso, Ridge, and elastic-net regressions. Instead of solely minimizing the average cross-entropy, these methods involve minimizing a modified objective function that includes a penalty term. Regularization is particularly beneficial for addressing common issues such as multicollinearity and overfitting. 

\paragraph{Log-linear Poisson regression.}

In the Poisson regression context, one notable specification for the regression function is the log-linear form, where the mean parameter is expressed as the exponential of a linear function of the predictors:
\begin{align}
    \mu^\text{LL}(\boldsymbol{x}; \boldsymbol{\beta}) = \exp\left\{\langle\boldsymbol{x}, \boldsymbol{\beta}\rangle\right\},
    \label{eq:log-linear}
\end{align}
where $\boldsymbol{\beta}$ denotes a vector of parameters, and $\langle\boldsymbol{x}, \boldsymbol{\beta}\rangle$ stands for the inner product between the predictor vector $\boldsymbol{x}$ and the coefficient vector $\boldsymbol{\beta}$. The use of the exponential function ensures that the mean parameter remains positive. Log-linear Poisson regression has favorable properties, notably its interpretability stemming from the quasi-linearity of the link function $\mu(\cdot)$. Moreover, when maximum likelihood is used for parameter estimation, this regression model falls within the framework of generalized linear models. GLMs provide valuable properties, such as the asymptotic Gaussian distribution of the parameters $\boldsymbol{\beta}$, allowing for the estimation of standard errors, hypothesis testing, and construction of confidence intervals.

However, log-linear regression does have a significant drawback -- its regression function, being linear, lacks flexibility. To address this limitation, various techniques can be employed. In fact, any supervised learning technique could be used for the specification of $\mu(\cdot)$. One simple approach to incorporating non-linearity involves adding polynomial terms of the predictors to the model alongside the linear terms. Splines, on the other hand, offer a flexible and powerful method for modeling non-linear relationships. Instead of fitting a single global function, splines divide the predictor range into smaller intervals and fit separate polynomial functions within each interval. This approach enables more localized and flexible modeling of the relationship between the predictors and the mean parameter.

\paragraph{CANN Poisson regression.}

In some cases, the supervised learning problem may require even more flexibility, and neural networks are particularly useful in such scenarios. Neural networks are formidable function approximation machines, well-known for their ability to estimate a wide range of highly non-linear multivariate functions. One of the key advantages of neural networks is their ability to handle raw and unstructured data effectively. Because we deal with detailed telematics data, this capability forms the basis for adopting the Combined Actuarial Neural Network (CANN) approach of \cite{wuthrich2019yes}, which embeds a classical actuarial model into a neural network architecture. A CANN model consists of two distinct components: the classical regression model component and the neural network component. This architecture offers great flexibility, allowing for seamless integration of any classical model whose regression function is compatible with a neural network architecture. Likewise, the neural network component can employ various types of supervised architectures, such as convolutional neural networks (CNNs), recurrent neural networks (RNNs), and other architectures tailored to the specific problem at hand. The classical regression model provides good initial estimations and serves as a guide for the neural network component. It offers a starting point for the network's optimization process, enabling faster convergence. The neural network component, in turn, refines the initial estimations, capturing additional signals and uncovering patterns that may have been missed by the classical model alone.

In our specific case, we use log-linear count regression as the classical model and a multilayer perceptron (MLP) as the neural network component in the CANN model. As a result, we have the following specification for the regression function:
\begin{align}
    \mu^\text{CANN}(\boldsymbol{x}; \boldsymbol{\beta}, \boldsymbol{\theta}) &=  \mu^\text{LL}(\boldsymbol{x}; \boldsymbol{\beta}) \times \mu^\text{MLP}(\boldsymbol{x}; \boldsymbol{\theta}),
    \label{eq:cann_spec_1}
\end{align}
where $\mu^\text{MLP}(\cdot)$ is the regression function learned by a multilayer perceptron parametrized with $\boldsymbol{\theta}$. In a nutshell, an MLP consists of interconnected layers, including an input layer, hidden layers, and an output layer. Each layer applies an affine transformation to the inputs it receives, followed by a non-linear activation function. This combination of linear transformations and non-linear activations allows MLPs to model complex non-linear relationships in the data. To delve into the mathematical description of an MLP, we can break down its structure starting from the input layer and progressing towards the output layer:
\begin{enumerate}
    \item \textbf{Input layer ($l = 0$)}:
    The input layer consists of $n_0$ nodes representing the input variables $\boldsymbol{x} = [x_1, x_2, \dots, x_{n_0}]$.
    \item \textbf{First hidden layer ($l = 1$):}
    The first hidden layer contains $n_1$ nodes, connected to the nodes from the input layer ($l = 0$) and the nodes in the subsequent layer ($l = 2$). The computations in the first hidden layer involve an affine transformation of the input variables followed by the application of a non-linear activation function, which introduces non-linearity into the network. Let us denote the weight matrix between layers $l = 0$ and $l = 1$ as $\boldsymbol{W}^{(1)}$ with dimensions $(n_1, n_0)$ and the bias vector as $\boldsymbol{b}^{(1)}$ with dimensions $(n_1, 1)$. The activation function applied to the transformed inputs is denoted as $\phi$. The computations in the first hidden layer can then be expressed as:
    \begin{align}
        \boldsymbol{a}^{(1)} = \boldsymbol{W}^{(1)} \boldsymbol{x} + \boldsymbol{b}^{(1)}, \quad \boldsymbol{z}^{(1)} = \phi\left( \boldsymbol{a}^{(1)}\right),
    \end{align}
    where $\boldsymbol{a}^{(1)}$ represents the preactivation values in the first hidden layer, and $\boldsymbol{z}^{(1)}$ represents the post-activation values. It is worth noting that the activation function $\phi$ is applied element-wise on the preactivation vector $\boldsymbol{a}^{(1)}$.
    \item \textbf{Subsequent hidden layers ($l = 2, 3, \dots, L - 2$)}:
    Each subsequent hidden layer $l$ contains $n_l$ nodes, connected to the nodes from the previous layer ($l - 1$) and the nodes in the following layer ($l + 1$). Similar to the first layer, the computations in the subsequent hidden layers involve an affine transformation of the inputs $\boldsymbol{z}^{(l-1)}$ followed by the application of the non-linear activation function $\phi$. Let us denote the weight matrix between layers $l - 1$ and $l$ as $\boldsymbol{W}^{(l)}$ with dimensions $(n_l, n_{l-1})$ and the bias vector as $\boldsymbol{b}^{(l)}$ with dimensions $(n_l, 1)$. The computations in the hidden layers can then be expressed as:
    \begin{align}
        \boldsymbol{a}^{(l)} = \boldsymbol{W}^{(l)} \boldsymbol{z}^{(l - 1)} + \boldsymbol{b}^{(l)}, \quad \boldsymbol{z}^{(l)} = \phi\left( \boldsymbol{a}^{(l)}\right),
    \end{align}
    where $\boldsymbol{a}^{(l)}$ represents the preactivation values in the $l^\text{th}$ hidden layer, and $\boldsymbol{z}^{(l)}$ represents the post-activation values.
    \item \textbf{Output layer ($l = L - 1$)}:
    The output layer consists of $n_{L-1}$ nodes, representing the final output(s) of the MLP. Similar to the hidden layers, the output layer involves an affine transformation followed by an activation function $g$. We denote the weight matrix between layers $L - 2$ (last hidden layer) and $L-1$ as $\boldsymbol{W}^{(L-1)}$ with dimensions $(n_{L-1}, n_{L-2})$ and the bias vector as $\boldsymbol{b}^{(L-1)}$ with dimensions $(n_{L-1}, 1)$. The computations within the output layer can be expressed as:
    \begin{align}
        \boldsymbol{a}^{(L - 1)} = \boldsymbol{W}^{(L - 1)} \boldsymbol{z}^{(L - 2)} + \boldsymbol{b}^{(L - 1)}, \quad \boldsymbol{z}^{(L - 1)} = g\left( \boldsymbol{a}^{(L - 1)}\right).
    \end{align}
\end{enumerate}
Note that the number of output neurons $n_{L-1}$ should match the number of modeled distribution parameters. In the context of Poisson regression, where we are modeling a single parameter $\mu$, only one output neuron is necessary. The choice of the output activation function $g(\cdot)$ is important and should be aligned with the specific problem being tackled since it determines the range and properties of the output values. For instance, in the classic case of a multi-class classification problem (where the multinoulli distribution is used as a specification for the target variable), each output neuron represents a class, and the predicted probabilities for each class should be positive and sum up to 1. In this scenario, a common choice for the activation function is the softmax function, which normalizes the outputs and ensures they are positive and sum up to 1. In our case, we need to ensure that the parameter $\mu$, which represents the expected count, is always positive. While the exponential function is a natural choice to enforce positivity, it can sometimes lead to numerical instability, especially for large input values. As a better alternative, we choose to use the softplus function as the activation function for the output layer, defined as $\zeta(x) = \log(1 + \exp(x))$. The softplus function is well-behaved even for large input values, mitigating the issue of numerical instability that can arise with the exponential function. The parameters of the generic MLP described above, consisting of weight matrices and bias vectors, can be denoted as:
\begin{align}
    \boldsymbol{\theta} = \left\{ \boldsymbol{W}^{(1)}, \boldsymbol{b}^{(1)}, \boldsymbol{W}^{(2)}, \boldsymbol{b}^{(2)}, \ldots, \boldsymbol{W}^{(L-1)}, \boldsymbol{b}^{(L-1)} \right\}.
\end{align}

\begin{figure}[ht]
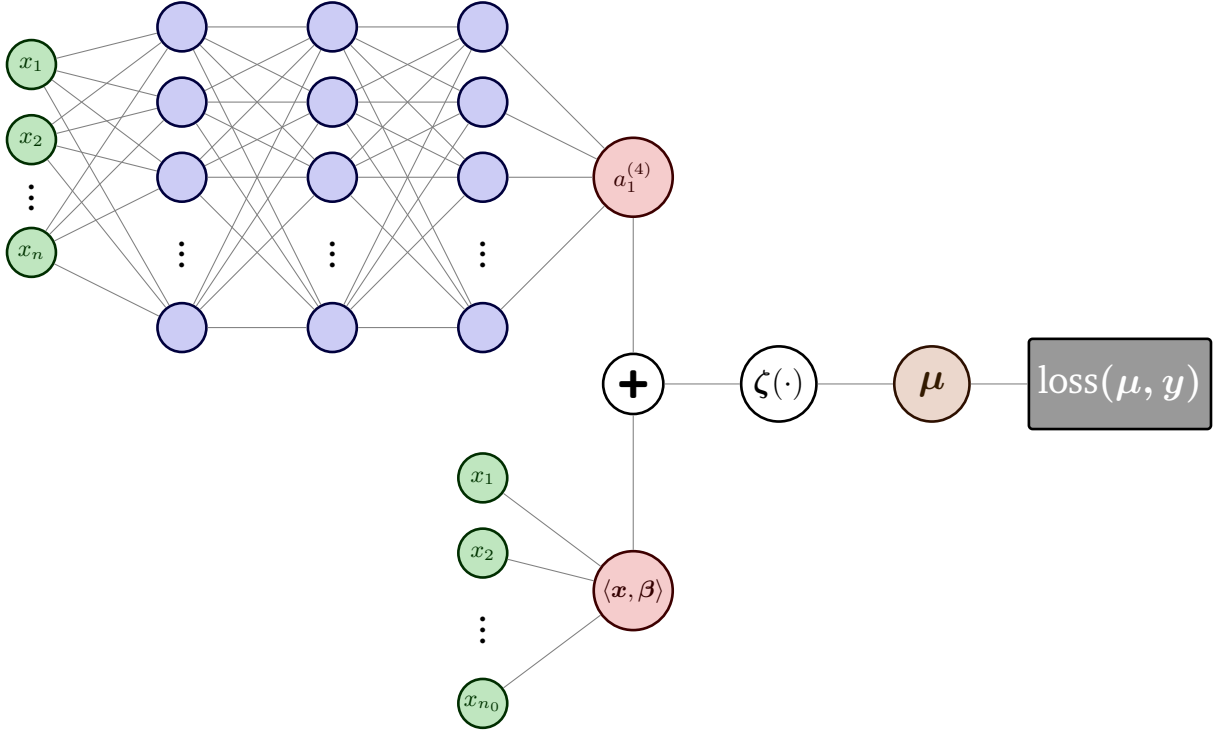

    \centering
    
    \tikz[every node/.style={rounded corners=.5mm}, font=\small]{
        \tikzstyle{circ}=[circle, inner sep=0pt, minimum size=.65cm, draw, line width=1pt]
        \tikzstyle{input}=[circle, inner sep=0pt, minimum size=.65cm, line width=1pt,green!20!black,draw=mygreen!30!black,fill=mygreen!25]
        \tikzstyle{hidden}=[circle, inner sep=0pt, minimum size=.65cm, line width=1pt,blue!20!black,draw=myblue!30!black,fill=myblue!20]
        \tikzstyle{output}=[circle, inner sep=0pt, minimum size=.65cm, line width=1pt,red!20!black,draw=myred!30!black,fill=myred!20]
        \tikzstyle{oran}=[circle, inner sep=0pt, minimum size=.65cm, line width=1pt,orange!20!black,draw=myorange!30!black,fill=myorange!20]

        \foreach \y[count=\i] in {0,1,2,4}{
            \foreach \x[count=\j] in {0,2,4}
                \node[hidden] (h\j-\i) at (\x,-\y) {};
        }
        
        \node[input, xshift=-2cm] (i1) at ($(h1-1)!.5!(h1-2)$) {$x_1$};
        \node[input, xshift=-2cm] (i2) at ($(h1-2)!.5!(h1-3)$) {$x_2$};
        \node[input, xshift=-2cm] (i3) at ($(h1-3)!.5!(h1-4)$) {$x_n$};
        
        \node[output, xshift=2cm, minimum size=1.05cm] (o1) at (h3-3) {$a_1^{(4)}$};
        
        \foreach \i in {1,2,3}{
            \foreach \j in {1,2,3,4}
                \draw[gray] (i\i)--(h1-\j);
        }
        \foreach \i in {1,2,3,4}{
            \foreach \j in {1,2,3,4}
                \draw[gray] (h1-\i)--(h2-\j) (h2-\i)--(h3-\j);
        }
        \foreach \i in {1,2,3,4}{
            \draw[gray] (h3-\i)--(o1);
        }
    
        \node[input, yshift=-2cm] (x1) at (h3-4) {$x_1$};
        \node[input, yshift=-1cm] (x2) at (x1) {$x_2$};
        \node[input, yshift=-2cm] (xn) at (x2) {$x_{n_0}$};
        \node[output, yshift=-.5cm, minimum size=1.05cm] (o2) at (o1|-x2) {$\langle \boldsymbol{x},\boldsymbol{\beta}\rangle$};
        
        \node[circ,  minimum size=0.8cm] (plus) at ($(o1)!0.5!(o2)$) {\large \faPlus};
        \node[circ,  minimum size=1cm, right=1cm of plus] (zeta)  {\large $\boldsymbol{\zeta}(\cdot)$};
        \node[oran,  minimum size=1cm, right=1cm of zeta] (mu)  {\Large $\boldsymbol{\mu}$};
        \node[draw, text=white, fill=mygray, minimum width=2cm, minimum height=1.2cm, xshift=2.5cm, line width=1pt] (l) at (mu) {\Large $\boldsymbol{\text{loss}(\mu,y)}$};
        \node[] at ($(i2)!.5!(i3)$) {\Large\strut$\vdots$};
         \foreach \i in {1,2,3}
        \node[] at ($(h\i-3)!.5!(h\i-4)$) {\Large\strut$\vdots$};
        \node[] at ($(x2)!.5!(xn)$) {\Large\strut$\vdots$};
        \foreach \i in {1,2,n}{
        \draw[gray] (x\i)--(o2);
        }
        
        \draw[gray] (o1)--(plus) (plus)--(o2) (plus)--(zeta) (zeta)--(mu) (mu)--(l);
    }
    \caption{CANN architecture for the Poisson specification. The MLP's preactivation output value $a_1^{(4)}$ is added to the log-linear model's preactivation output value $\langle \boldsymbol{x}, \boldsymbol{\beta} \rangle$ before being transformed with the softplus function $\zeta(\cdot)$. The resulting $\mu$ value is compared to the ground truth $y$ using Poisson cross-entropy loss. The architecture shown employs a 3-hidden-layer MLP, but can be customized with any number of layers.}
    \label{fig:CG_poisson}
\end{figure}

Naturally, these parameters must be estimated. However, the criterion in Equation~(\ref{eq:loss_poisson}) is typically not convex, making it challenging to find the global minimum of the empirical risk. In practice, the goal is to find a ``good enough'' local minimum that yields satisfactory performance on the task. Gradient descent algorithms, such as stochastic gradient descent (SGD) and its variants, are commonly employed to update iteratively the parameters $\boldsymbol{\theta}$ in the direction of the steepest descent. The backpropagation algorithm efficiently computes the gradients and propagates them through the network, enabling parameter updates. With the introduced notation for the MLP, we can now express the specification in (\ref{eq:cann_spec_1}) as
\begin{align}
    \mu^\text{CANN}(\boldsymbol{x}; \boldsymbol{\beta}, \boldsymbol{\theta}) &= \zeta\left\{\langle\boldsymbol{x}, \boldsymbol{\beta}\rangle + \boldsymbol{a}^{(L - 1)}(\boldsymbol{x}; \boldsymbol{\theta})\right\}.
    \label{eq:spec_mu_poisson}
\end{align}
The corresponding computational graph for $L = 5$ (i.e., 3 hidden layers) is shown in Figure~\ref{fig:CG_poisson}. 
Like a standard MLP, the network parameters $\boldsymbol{\beta}$ and $\boldsymbol{\theta}$ can be estimated using gradient descent. A simplified pseudo-algorithm for the training of the Poisson CANN model is provided in Algorithm~\ref{alg:gradient_descent_poisson}. The learning rate $\eta$ is a hyperparameter that determines the step size taken every time a gradient descent step is performed. In other words, it controls how quickly or slowly the network parameters are updated during training. A higher learning rate allows for larger steps, which can lead to faster convergence. However, an excessively high learning rate may cause the optimization process to overshoot or oscillate around the minimum, hindering convergence. Conversely, a very low learning rate might result in slow convergence, requiring more iterations to reach an acceptable solution. Finding the right learning rate is important and is typically an empirical process that requires experimentation and tuning.
\begin{algorithm}[ht]
    \SetAlgoLined
    \KwIn{Training dataset $\{(\boldsymbol{x}_{it}, y_{it})\}_{(i, t) \in \mathcal{T}_r}$, learning rate $\eta$, number of epochs $E$}
    \KwOut{Trained model parameters $\hat{\boldsymbol{\theta}}$, $\hat{\boldsymbol{\beta}}$ and $\hat{w}_\phi$}
    \BlankLine
    Initialize model parameters $\hat{\boldsymbol{\theta}} = \boldsymbol{0}$, $\hat{\boldsymbol{\beta}} = \hat{\boldsymbol{\beta}}^\text{MLE}$ and $\hat{w}_\phi$.\\
    \BlankLine
    \For{epoch $\leftarrow$ 1 \KwTo $E$}{
         \begin{enumerate}
            \item For each contract $(i, t)$, apply the CANN regression function with current network parameters to compute the current estimated mean parameter $\hat{\mu}_{it}$:
            \begin{align*}
                \hat{\mu}_{it} = \mu^\text{CANN}(\boldsymbol{x}_{it}; \hat{\boldsymbol{\beta}}, \hat{\boldsymbol{\theta}}).
            \end{align*}
            \item Compute the empirical risk over the training dataset: 
            \begin{align*}
                \mathcal{R} = -\frac{1}{|\mathcal{T}_r|} \sum_{(i, t) \in \mathcal{T}_r} y_{it} \ln[\hat{\mu}_{it}] - \hat{\mu}_{it} - y_{it}!.
            \end{align*}
            \item Perform backpropagation to compute the gradients of $\mathcal{R}$ with respect to the network parameters: 
            \begin{align*}
                \nabla_{\boldsymbol{\beta}} \mathcal{R} \quad\text{and}\quad \nabla_{\boldsymbol{\theta}} \mathcal{R}.
            \end{align*}
            \item  Perform gradient descent using the learning rate:
            \begin{align*}
                \hat{\boldsymbol{\beta}} \leftarrow \hat{\boldsymbol{\beta}} - \eta \nabla_{\hat{\boldsymbol{\beta}}} \mathcal{R},\\
                \hat{\boldsymbol{\theta}} \leftarrow \hat{\boldsymbol{\theta}} - \eta \nabla_{\hat{\boldsymbol{\theta}}} \mathcal{R}.
            \end{align*}
        \end{enumerate}
    }
    \caption{Parameter estimation procedure -- Poisson CANN model}
    \label{alg:gradient_descent_poisson}
\end{algorithm}
In practice, mini-batch gradient descent is commonly used for training neural networks. It works by dividing the training data into smaller subsets, called mini-batches, and computing the gradients and parameter updates based on these mini-batches. This approach offers computational efficiency and improved generalization compared to regular gradient descent, making it a preferred choice in practice. For a comprehensive understanding of neural networks, we refer to the excellent book \cite{goodfellow2016deep}.

\subsubsection{Negative binomial regression}

One issue with the Poisson distribution is its equidispersion assumption. Indeed, we have that $\mu(\boldsymbol{x}) = \Esp{Y_{it}|\boldsymbol{X} = \boldsymbol{x}} = \text{Var}[{Y_{it}|\boldsymbol{X} = \boldsymbol{x}}]$. In practice, claim count data often exhibit overdispersion, where the observed variance of the claim count is greater than the mean. To address this limitation, alternative distributions allowing for overdispersion can be used. Among them, the negative binomial distribution (see, for instance, \cite{denuit2007actuarial} and \cite{cameron2013regression}) stands out as a common choice. Under the negative binomial specification, the PMF of the claim count for the $t^\text{th}$ contract of vehicle $i$ ($Y_{it}$), given its predictor vector ($\boldsymbol{x}_{it}$), can be written as
\begin{align}
    \mathbb{P}(Y_{it} = y_{it}|\boldsymbol{x}_{it}) = \frac{\Gamma(y_{it} + \phi)}{y_{it}! \Gamma(\phi)} \left(\frac{\phi}{\phi + \mu(\boldsymbol{x}_{it})}\right)^\phi \left( \frac{\mu(\boldsymbol{x}_{it})}{\mu(\boldsymbol{x}_{it}) + \phi}\right)^{y_{it}},\quad \text{for} \quad y_{it} \in \mathbb{N},
    \label{eq:nb2}
\end{align}
where $\phi > 0$ is a dispersion parameter. This can be seen as a generalization of the Poisson distribution. Indeed, the Poisson distribution is recovered when $\frac{1}{\phi} \rightarrow 0$. The first two centered moments are given by:
\begin{align}
    \Esp{Y_{it}|\boldsymbol{X}_{it} = \boldsymbol{x}_{it}} = \mu(\boldsymbol{x}_{it}) \quad \text{and} \quad \text{Var}[Y_{it}|\boldsymbol{X}_{it} = \boldsymbol{x}_{it}] = \mu(\boldsymbol{x}_{it}) + \frac{\mu(\boldsymbol{x}_{it})^2}{\phi}.
\end{align}
As can be seen, the negative binomial specification assumes overdispersion since $\text{Var}[Y_{it}|\boldsymbol{X}_{it} = \boldsymbol{x}_{it}] > \Esp{Y_{it}|\boldsymbol{X}_{it} = \boldsymbol{x}_{it}}$. Once the specification for the regression function $\mu(\cdot)$ has been chosen, which defines a set of candidate functions $\mathcal{H}$, one can estimate the parameters of the regression function $\mu(\cdot)$ along with the dispersion parameter $\phi$ by maximum likelihood or, equivalently, by minimizing the empirical risk over the training set:
\begin{align}
     \{\widehat{\mu}, \widehat{\phi}\} = \argmin_{\mu \in \mathcal{H}, \phi > 0} \left\{ -\frac{1}{|\mathcal{T}_r|}\sum_{(i,t) \in \mathcal{T}_r} \ln \left[\frac{\Gamma(y_{it} + \phi)}{y_{it}!\Gamma(\phi)}\right] + \phi \ln\left[\frac{\phi}{\phi + \mu(\boldsymbol{x}_{it})}\right] + y_{it} \ln\left[\frac{\mu(\boldsymbol{x}_{it})}{\mu(\boldsymbol{x}_{it}) + \phi}\right]\right\}.
     \label{eq:emp_risk_nb2}
\end{align}

\paragraph{Log-linear negative binomial regression.}

As in the Poisson case, a common specification for $\mu(\cdot)$ is the log-linear form, defined in Equation~(\ref{eq:log-linear}). In this case, the criterion in (\ref{eq:emp_risk_nb2}) is convex, and convex optimization can be used to estimate $\boldsymbol{\beta}$ and $\phi$.

\paragraph{CANN negative binomial regression.}

\begin{figure}[ht]
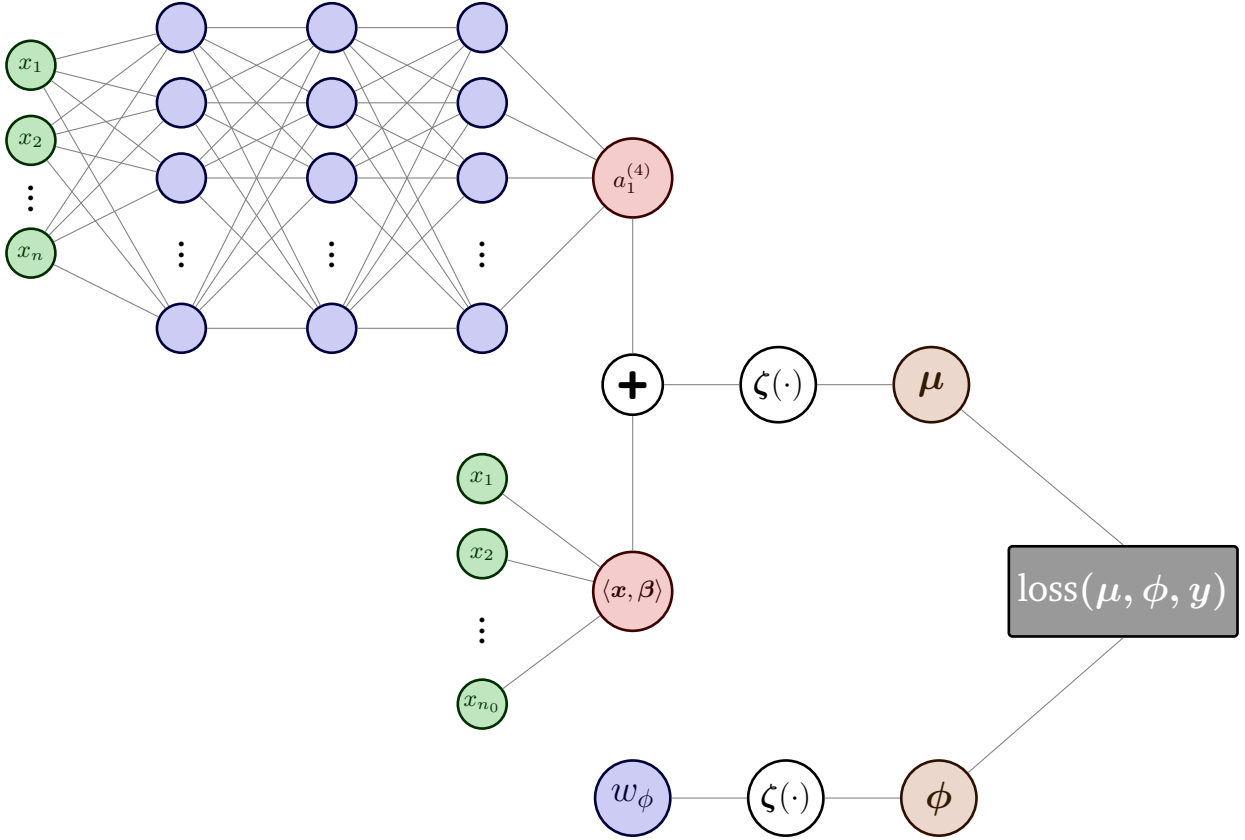

    \centering
    
    \tikz[every node/.style={rounded corners=.5mm}, font=\small]{
        \tikzstyle{circ}=[circle, inner sep=0pt, minimum size=.65cm, draw, line width=1pt]
        \tikzstyle{input}=[circle, inner sep=0pt, minimum size=.65cm, line width=1pt,green!20!black,draw=mygreen!30!black,fill=mygreen!25]
        \tikzstyle{hidden}=[circle, inner sep=0pt, minimum size=.65cm, line width=1pt,blue!20!black,draw=myblue!30!black,fill=myblue!20]
        \tikzstyle{output}=[circle, inner sep=0pt, minimum size=.65cm, line width=1pt,red!20!black,draw=myred!30!black,fill=myred!20]
        \tikzstyle{oran}=[circle, inner sep=0pt, minimum size=.65cm, line width=1pt,orange!20!black,draw=myorange!30!black,fill=myorange!20]

        \foreach \y[count=\i] in {0,1,2,4}{
            \foreach \x[count=\j] in {0,2,4}
                \node[hidden] (h\j-\i) at (\x,-\y) {};
        }
        
        \node[input, xshift=-2cm] (i1) at ($(h1-1)!.5!(h1-2)$) {$x_1$};
        \node[input, xshift=-2cm] (i2) at ($(h1-2)!.5!(h1-3)$) {$x_2$};
        \node[input, xshift=-2cm] (i3) at ($(h1-3)!.5!(h1-4)$) {$x_n$};
        
        \node[output, xshift=2cm, minimum size=1.05cm] (o1) at (h3-3) {$a_1^{(4)}$};
        
        \foreach \i in {1,2,3}{
            \foreach \j in {1,2,3,4}
                \draw[gray] (i\i)--(h1-\j);
        }
        \foreach \i in {1,2,3,4}{
            \foreach \j in {1,2,3,4}
                \draw[gray] (h1-\i)--(h2-\j) (h2-\i)--(h3-\j);
        }
        \foreach \i in {1,2,3,4}{
            \draw[gray] (h3-\i)--(o1);
        }
    
        \node[input, yshift=-2cm] (x1) at (h3-4) {$x_1$};
        \node[input, yshift=-1cm] (x2) at (x1) {$x_2$};
        \node[input, yshift=-2cm] (xn) at (x2) {$x_{n_0}$};
        \node[output, yshift=-.5cm, minimum size=1.05cm] (o2) at (o1|-x2) {$\langle \boldsymbol{x},\boldsymbol{\beta}\rangle$};
        
        \node[circ,  minimum size=0.8cm] (plus) at ($(o1)!0.5!(o2)$) {\large \faPlus};
        \node[circ,  minimum size=1cm, right=1cm of plus] (phi)  {\large $\boldsymbol{\zeta}(\cdot)$};
        \node[oran,  minimum size=1cm, right=1cm of phi] (mu)  {\Large $\boldsymbol{\mu}$};
        \node[hidden,  minimum size=1cm] (w) at ([yshift=-2.75cm]o2) {\Large $w_\phi$};
        \node[circ,  minimum size=1cm, right=1cm of w] (phi2)  {\large $\boldsymbol{\zeta}(\cdot)$};
        \node[oran,  minimum size=1cm, right=1cm of phi2] (alpha)  {\Large $\boldsymbol{\phi}$};
        \node[draw, text=white, fill=mygray, minimum width=2cm, minimum height=1.2cm, xshift=2.5cm, line width=1pt] (l) at ($(mu)!0.5!(alpha)$) {\Large $\boldsymbol{\text{loss}(\mu,\phi,y)}$};
        \node[] at ($(i2)!.5!(i3)$) {\Large\strut$\vdots$};
         \foreach \i in {1,2,3}
        \node[] at ($(h\i-3)!.5!(h\i-4)$) {\Large\strut$\vdots$};
        \node[] at ($(x2)!.5!(xn)$) {\Large\strut$\vdots$};
        \foreach \i in {1,2,n}{
        \draw[gray] (x\i)--(o2);
        }
        
        \draw[gray] (o1)--(plus) (plus)--(o2) (plus)--(phi) (phi)--(mu) (mu)--(l.north) (l.south)--(alpha) (alpha)--(phi2) (phi2)--(w);
    }
    \caption{CANN architecture for the negative binomial specification. The MLP's preactivation output value $a_1^{(4)}$ is added to the log-linear model's preactivation output value $\langle \boldsymbol{x}, \boldsymbol{\beta} \rangle$ before being transformed with the softplus function $\zeta(\cdot)$ to obtain the $\mu$ value of the negative binomial distribution. The $\phi$ value is obtained by transforming a real-valued parameter $w_\phi$ through the softplus function. The resulting parameters $\mu$ and $\phi$ are then compared to the ground truth $y$ using negative binomial cross-entropy loss. The architecture shown employs a 3-hidden-layer MLP, but can be customized with any number of layers.}
    \label{fig:CG_nb2}
\end{figure}

Similar to the approach used for the Poisson case, a CANN architecture can be used to model the mean parameter in the negative binomial distribution. The specification for the regression function $\mu(\cdot)$ remains identical to the Poisson case, as defined in Equation~(\ref{eq:spec_mu_poisson}). In order to incorporate the extra distribution parameter $\phi$, an additional output neuron is introduced in the network. This output neuron is connected to a neural network weight $w_\phi \in \mathbb{R}$ through the softplus function, ensuring that $\phi$ remains positive, i.e., $\phi = \zeta(w_\phi)$. It is important to highlight that the distribution parameter $\phi$ is not directly connected to the input variables $\boldsymbol{x}$. As a result, no heterogeneity is incorporated into this parameter, and a common estimated value $\widehat{\phi}$ is used for all observations. The exact architecture for the negative binomial CANN model is depicted in Figure~\ref{fig:CG_nb2}.
The network parameters $\boldsymbol{\beta}$, $\boldsymbol{\theta}$, and $w_\phi$ can be learned by minimizing the criterion in Equation~(\ref{eq:emp_risk_nb2}) using the procedure described in Algorithm~\ref{alg:gradient_descent_nb2}.
\begin{algorithm}[ht]
    \SetAlgoLined
    \KwIn{Training dataset $\{(\boldsymbol{x}_{it}, y_{it})\}_{(i, t) \in \mathcal{T}_r}$, learning rate $\eta$, number of epochs $E$}
    \KwOut{Trained model parameters $\hat{\boldsymbol{\theta}}$, $\hat{\boldsymbol{\beta}}$ and $\hat{w}_\phi$}
    \BlankLine
    Initialize model parameters $\hat{\boldsymbol{\theta}} = \boldsymbol{0}$, $\hat{\boldsymbol{\beta}} = \hat{\boldsymbol{\beta}}^\text{MLE}$ and $\hat{w}_\phi$.\\
    \BlankLine
    \For{epoch $\leftarrow$ 1 \KwTo $E$}{
        \begin{enumerate}
            \item For each contract $(i, t)$, apply the CANN regression function with current network parameters to compute the current estimated mean parameter $\hat{\mu}_{it}$:
            \begin{align*}
                \hat{\mu}_{it} = \mu^\text{CANN}(\boldsymbol{x}_{it}; \hat{\boldsymbol{\beta}}, \hat{\boldsymbol{\theta}}).
            \end{align*}
            \item Compute the current estimated parameter $\hat{\phi}$:
            \begin{align*}
                \hat{\phi} = \zeta(\hat{w}_\phi).
            \end{align*}
            \item Compute the empirical risk over the training dataset: 
            \begin{align*}
                \mathcal{R} = -\frac{1}{|\mathcal{T}_r|}\sum_{(i,t) \in \mathcal{T}_r} \ln \left[\frac{\Gamma(y_{it} + \hat{\phi})}{y_{it}!\Gamma(\hat{\phi})}\right] + \hat{\phi} \ln\left[\frac{\hat{\phi}}{\hat{\phi} + \hat{\mu}_{it}}\right] + y_{it} \ln\left[\frac{\hat{\mu}_{it}}{\hat{\mu}_{it} + \hat{\phi}}\right].
            \end{align*}
            \item Perform backpropagation to compute the gradients of $\mathcal{R}$ with respect to the network parameters: 
            \begin{align*}
                \nabla_{\boldsymbol{\beta}} \mathcal{R} \text{,}\quad \nabla_{\boldsymbol{\theta}} \mathcal{R} \quad\text{and}\quad \nabla_{w_\phi} \mathcal{R}.
            \end{align*}
            \item  Perform gradient descent using the learning rate:
            \begin{align*}
                \hat{\boldsymbol{\beta}} \leftarrow \hat{\boldsymbol{\beta}} - \eta \nabla_{\hat{\boldsymbol{\beta}}} \mathcal{R},\\
                \hat{\boldsymbol{\theta}} \leftarrow \hat{\boldsymbol{\theta}} - \eta \nabla_{\hat{\boldsymbol{\theta}}} \mathcal{R},\\
                \hat{w}_\phi \leftarrow \hat{w}_\phi - \eta \nabla_{\hat{w}_\phi} \mathcal{R}.
            \end{align*}
        \end{enumerate}
    }
    \caption{Parameter estimation procedure -- Negative binomial CANN model}
    \label{alg:gradient_descent_nb2}
\end{algorithm}

\subsection{Longitudinal models}

Cross-sectional models assume independence between all contracts. However, in our case, the data exhibits clustering due to contracts being grouped by vehicle. While it is reasonable to assume independence between contracts from distinct vehicles, this assumption is less valid for contracts from the same vehicle. In reality, the claim counts of contracts within the same vehicle may be influenced by shared vehicle-specific characteristics, unobserved risk factors, or policy-level effects, resulting in dependence between observations within each vehicle cluster. To appropriately address this dependence, we transition from cross-sectional to longitudinal models, enabling the introduction of within-vehicle dependence. In the case of claim count data, a longitudinal model can efficiently leverage the history of the vehicles to refine the risk estimation for future contracts.

While various models are available to analyze longitudinal data, such as random effects models, fixed effects models, generalized estimating equations (GEE), and autoregressive models (AR), among others, empirical evidence in the context of claim count regression supports the effectiveness of random (or mixed) effects models (see \cite{boucher2008models}). In these models, a random effect, which is a random variable, is introduced in the specified distribution. For instance, in the case of count data, the specified distribution could be the Poisson distribution. The random effect is assumed to follow a certain distribution, such as a normal, gamma, or another appropriate distribution. The inclusion of the random effect allows for capturing the unobserved heterogeneity or individual-specific effects that cannot be accounted for by the observed covariates. It introduces additional variability into the model and accounts for the dependence within clusters. In longitudinal analysis, we need, for each vehicle $i$, to model the random vector of claim counts $\boldsymbol{Y}_{i, (1:T_i)} = (Y_{i1}, \dots, Y_{i,T_i})$. The joint PMF can be expressed with
\begin{align}
    \mathbb{P}\left(\boldsymbol{Y}_{i, (1:T_i)} = \boldsymbol{y}_{i, (1:T_i)}| \boldsymbol{x}_{i, (1:T_i)}\right) = \int_{-\infty}^\infty \left(\prod_{t=1}^{T_i}\mathbb{P}(Y_{it} = y_{it}|\boldsymbol{x}_{i, (1:T_i)}, \theta_i)\right) f(\theta_i) d\theta_i,
    \label{eq:long_models}
\end{align}
where $f(\theta_i)$ is the PDF of the ramdom effect.

\subsubsection{Multivariate negative binomial regression}

A multivariate negative binomial regression model is obtained by introducing a gamma-distributed random effect in the mean parameter of the Poisson distribution. Specifically, we assume that the conditional distribution of $Y_{it}$, given $\Theta_i = \theta_i$, follows a Poisson distribution with mean $\mu(\boldsymbol{x}_{it})\theta_i$, where $\Theta_i$ is a gamma-distributed random variable with mean $1$ and variance $1/\phi$. The density of $\Theta_i$ is given by
\begin{align}
f_{\Theta_i}(\theta_i) = \frac{\phi^\phi}{\Gamma(\phi)} \theta_i^{\phi - 1} e^{-\phi\theta_i}, \quad \theta_i > 0.
\end{align}
By using Equation~(\ref{eq:long_models}), one can derive the joint distribution for the vector of claim counts:
\begin{align}
    \mathbb{P}\left(\boldsymbol{Y}_{i, (1:T_i)} = \boldsymbol{y}_{i, (1:T_i)}| \boldsymbol{x}_{i, (1:T_i)}\right) = \prod_{t=1}^{T_i} \left(\frac{\mu(\boldsymbol{x}_{it})^{y_{it}}}{y_{it!}}\right) \frac{\Gamma(y_{i\bullet} + \phi)}{\Gamma(\phi)} \left(\frac{\phi}{\mu_{i\bullet} + \phi}\right)^\phi \left(\frac{1}{\mu_{i\bullet} + \phi}\right)^{y_{it}},
\end{align}
where $\mu_{i\bullet} = \sum_{t = 1}^{T_i} \mu_{it}$ and $y_{i\bullet} = \sum_{t = 1}^{T_i} y_{it}$. This joint distribution is commonly referred to as the multivariate negative binomial (MVNB) or negative multinomial distribution. Note that the Poisson distribution is retrieved when $\frac{1}{\phi} \rightarrow 0$. Furthermore, given the past claim history denoted as $\boldsymbol{y}_{i, (1:t-1)} = (y_{i1}, \dots, y_{i,t-1})$ as well as current and past covariate vectors denoted as $\boldsymbol{x}_{i, (1:t)} = (\boldsymbol{x}_{i1}, \dots, \boldsymbol{x}_{it})$, one can show that the number of claims at time (or contract) $t$ follows a negative binomial distribution. The probability of observing $y_{it}$ claims at time $t$, given the past claim history as well as past and current covariate vectors, is thus expressed with
\begin{align}
    \mathbb{P}(Y_{it} = y_{it}| \boldsymbol{y}_{i, (1:t-1)}, \boldsymbol{x}_{i, 1:t}) = \frac{\Gamma(y_{it} + \alpha_{it})}{y_{it}! \Gamma(\alpha_{it})} \left(\frac{\gamma_{it}}{\gamma_{it} + \mu(\boldsymbol{x}_{it})}\right)^{\alpha_{it}} \left( \frac{\mu(\boldsymbol{x}_{it})}{\mu(\boldsymbol{x}_{it}) + \gamma_{it}}\right)^{y_{it}},\quad t = 1, 2, \dots, T_i,
    \label{eq:gen_negbin}
\end{align}
where $\alpha_{it} = \phi + \Sigma_{it}^{(y)}$ and $\gamma_{it} = \phi + \Sigma_{it}^{(\mu)}$. $\Sigma_{it}^{(y)} = \sum_{t' = 1}^{t-1} y_{it'}$ and $\Sigma_{it}^{(\mu)} = \sum_{t' = 1}^{t-1} \mu(\boldsymbol{x}_{it'})$ represent the number of past claims and the sum of past $\mu$ values for contract $(i, t)$, resepctively. In the special case when $t=1$, there is no past history and we set $\Sigma_{it}^{(y)} = \Sigma_{it}^{(\mu)} = 0$, which yields $\alpha_{i1} = \gamma_{i1} = \phi$. The expected claim count, given the past history, is given by:
\begin{align}
    \Esp{Y_{it}| \boldsymbol{y}_{i, (1:t-1)}, \boldsymbol{x}_{i, 1:t}} &= \mu(\boldsymbol{x}_{it})\left(\frac{\phi + \Sigma_{it}^{(y)}}{\phi + \Sigma_{it}^{(\mu)}}\right)\\
    &= \mu(\boldsymbol{x}_{it})\left(\frac{\alpha_{it}}{\gamma_{it}}\right).
    \label{eq:esp_mvnb}
\end{align}
Fitting an MVNB model, therefore, amounts to fitting a negative binomial model, where the parameters $\alpha_{it}$ and $\gamma_{it}$ depend on the vehicle's history. Once the specification for the regression function $\mu(\cdot)$ is chosen, the parameter $\phi$ and the parameters in the regression function $\mu(\cdot)$ can be estimated by minimizing the empirical risk over the training set. This can be achieved through the following optimization problem:
\begin{align}
     \{\widehat{\mu}, \widehat{\phi}\} = \argmin_{\mu \in \mathcal{H}, \phi > 0} \left\{ -\frac{1}{|\mathcal{T}_r|}\sum_{(i,t) \in \mathcal{T}_r} \ln \left[\frac{\Gamma(y_{it} + \alpha_{it})}{y_{it}!\Gamma(\alpha_{it})}\right] + \alpha_{it} \ln\left[\frac{\gamma_{it}}{\gamma_{it} + \mu(\boldsymbol{x}_{it})}\right] + y_{it} \ln\left[\frac{\mu(\boldsymbol{x}_{it})}{\mu(\boldsymbol{x}_{it}) + \gamma_{it}}\right]\right\}.
     \label{eq:emp_risk_mvnb}
\end{align}

\paragraph{Log-linear multivariate negative binomial regression.}

If the specification for $\mu(\cdot)$ is the log-linear form, defined in Equation~(\ref{eq:log-linear}), the criterion in (\ref{eq:emp_risk_mvnb}) is convex, and convex optimization can be used to estimate $\boldsymbol{\beta}$ and $\phi$.

\paragraph{CANN multivariate negative binomial regression.}

\begin{figure}[ht]
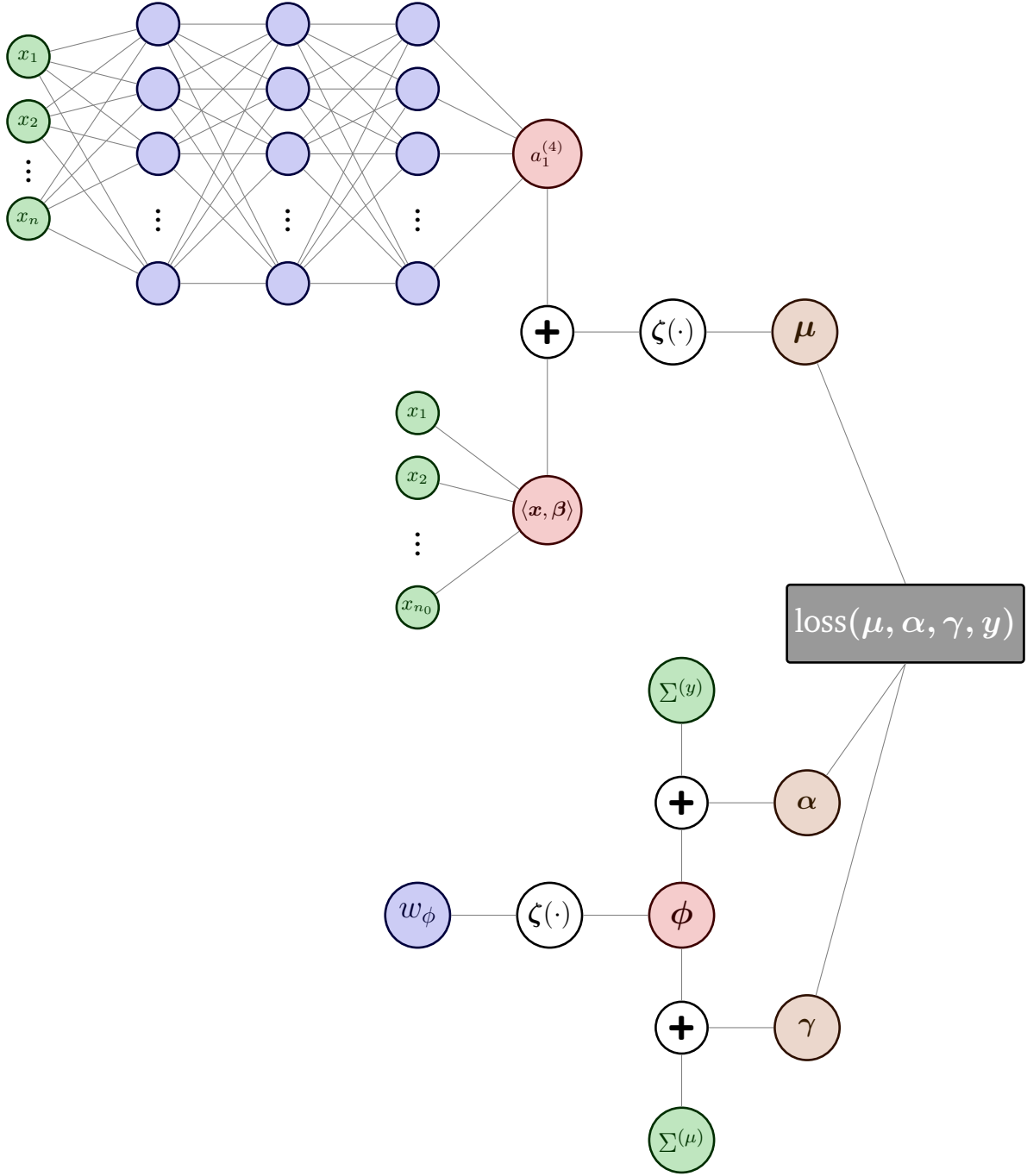

    \centering
    
    \tikz[every node/.style={rounded corners=.5mm}, font=\small]{
        \tikzstyle{circ}=[circle, inner sep=0pt, minimum size=.65cm, draw, line width=1pt]
        \tikzstyle{input}=[circle, inner sep=0pt, minimum size=.65cm, line width=1pt,green!20!black,draw=mygreen!30!black,fill=mygreen!25]
        \tikzstyle{hidden}=[circle, inner sep=0pt, minimum size=.65cm, line width=1pt,blue!20!black,draw=myblue!30!black,fill=myblue!20]
        \tikzstyle{output}=[circle, inner sep=0pt, minimum size=.65cm, line width=1pt,red!20!black,draw=myred!30!black,fill=myred!20]
        \tikzstyle{oran}=[circle, inner sep=0pt, minimum size=.65cm, line width=1pt,orange!20!black,draw=myorange!30!black,fill=myorange!20]

        \foreach \y[count=\i] in {0,1,2,4}{
            \foreach \x[count=\j] in {0,2,4}
                \node[hidden] (h\j-\i) at (\x,-\y) {};
        }
        
        \node[input, xshift=-2cm] (i1) at ($(h1-1)!.5!(h1-2)$) {$x_1$};
        \node[input, xshift=-2cm] (i2) at ($(h1-2)!.5!(h1-3)$) {$x_2$};
        \node[input, xshift=-2cm] (i3) at ($(h1-3)!.5!(h1-4)$) {$x_n$};
        
        \node[output, xshift=2cm, minimum size=1.05cm] (o1) at (h3-3) {$a_1^{(4)}$};
        
        \foreach \i in {1,2,3}{
            \foreach \j in {1,2,3,4}
                \draw[gray] (i\i)--(h1-\j);
        }
        \foreach \i in {1,2,3,4}{
            \foreach \j in {1,2,3,4}
                \draw[gray] (h1-\i)--(h2-\j) (h2-\i)--(h3-\j);
        }
        \foreach \i in {1,2,3,4}{
            \draw[gray] (h3-\i)--(o1);
        }
    
        \node[input, yshift=-2cm] (x1) at (h3-4) {$x_1$};
        \node[input, yshift=-1cm] (x2) at (x1) {$x_2$};
        \node[input, yshift=-2cm] (xn) at (x2) {$x_{n_0}$};
        \node[output, yshift=-.5cm, minimum size=1.05cm] (o2) at (o1|-x2) {$\langle \boldsymbol{x},\boldsymbol{\beta}\rangle$};
        
        \node[circ, minimum size=0.8cm] (plus) at ($(o1)!0.5!(o2)$) {\large \faPlus};
        \node[circ, minimum size=1cm, right=1cm of plus] (zeta)  {\large $\boldsymbol{\zeta}(\cdot)$};
        \node[oran, minimum size=1cm, right=1cm of zeta] (mu)  {\Large $\boldsymbol{\mu}$};
        \node[hidden, minimum size=1cm] (w) at ([yshift=-4.75cm]xn) {\Large $w_\phi$};
        \node[circ, minimum size=1cm, right=1cm of w] (zeta2)  {\large $\boldsymbol{\zeta}(\cdot)$};
        \node[output, minimum size=1cm, right=1cm of zeta2] (phi)  {\Large $\boldsymbol{\phi}$};
        \node[circ, minimum size=0.8cm, above=0.8cm of phi] (plus2) {\large \faPlus};
        \node[circ, minimum size=0.8cm, below=0.8cm of phi] (plus3) {\large \faPlus};
        \node[input, minimum size=1cm, above=0.8cm of plus2] (sigma_y) {\large $\Sigma^{(y)}$};
        \node[input, minimum size=1cm, below=0.8cm of plus3] (sigma_mu) {\large $\Sigma^{(\mu)}$};
        \node[oran, minimum size=1cm, right=1cm of plus2] (alpha) {\large $\boldsymbol{\alpha}$};
        \node[oran, minimum size=1cm, right=1cm of plus3] (gamma) {\large $\boldsymbol{\gamma}$};
        \node[draw, text=white, fill=mygray, minimum width=2cm, minimum height=1.2cm, xshift=2.5cm, line width=1pt] (l) at ($(mu)!0.5!(phi)$) {\Large $\boldsymbol{\text{loss}(\mu,\alpha, \gamma, y)}$};
        \node[] at ($(i2)!.5!(i3)$) {\Large\strut$\vdots$};
         \foreach \i in {1,2,3}
        \node[] at ($(h\i-3)!.5!(h\i-4)$) {\Large\strut$\vdots$};
        \node[] at ($(x2)!.5!(xn)$) {\Large\strut$\vdots$};
        \foreach \i in {1,2,n}{
            \draw[gray] (x\i)--(o2);
        }
        
        \draw[gray] (o1)--(plus) (plus)--(o2) (plus)--(zeta) (zeta)--(mu) (mu)--(l.north) (phi)--(zeta2) (zeta2)--(w) (phi)--(plus2) (phi)--(plus3) (plus2)--(sigma_y) (plus3)--(sigma_mu) (plus2)--(alpha) (plus3)--(gamma) (alpha)--(l.south) (gamma)--(l.south);
    }
    \caption{CANN architecture for the MVNB specification. The MLP's preactivation output value $a_1^{(4)}$ is added to the log-linear model's preactivation output value $\langle \boldsymbol{x}, \boldsymbol{\beta} \rangle$ before being transformed with the softplus function $\zeta(\cdot)$ to obtain the $\mu$ value of the negative binomial distribution of Equation~(\ref{eq:gen_negbin}). The $\phi$ value is obtained by transforming a real-valued parameter $w_\phi$ through the softplus function. To obtain $\alpha$, the sum of past claims $\Sigma^{(y)}$ is added to the $\phi$ parameter, while for $\gamma$, the sum of past $\mu$ values $\Sigma^{(\mu)}$ is added to the same $\phi$ parameter. The resulting distribution parameters $\mu$, $\alpha$ and $\gamma$ are then compared to the ground truth $y$ using negative binomial cross-entropy loss. The architecture shown employs a 3-hidden-layer MLP, but can be customized with any number of layers.}
    \label{fig:CG_mvnb}
\end{figure}

In the MVNB case, the CANN architecture, as defined in Equation~(\ref{eq:spec_mu_poisson}), can also be used as a specification for the regression function $\mu(\cdot)$. To incorporate the additional distribution parameters $\alpha_{it}$ and $\gamma_{it}$, two additional output neurons are introduced in the network, as depicted in Figure~\ref{fig:CG_mvnb}. 
The distribution parameters $\alpha_{it}$ and $\gamma_{it}$ stem from a common parameter $\phi > 0$, and for the $t^\text{th}$ contract of vehicle $i$, we have $\alpha_{it} = \phi + \Sigma_{it}^{(y)}$ and $\gamma_{it} = \phi + \Sigma_{it}^{(\mu)}$. The neuron representing $\phi$ is connected to a network weight $w_\phi \in \mathbb{R}$ through the softplus function, i.e., $\phi = \zeta(w_\phi)$. An MVNB CANN model can be trained with backpropagation and gradient descent, as outlined in Algorithm~\ref{alg:gradient_descent_mvnb}. Notice that for a vehicle $i$, the parameter $\gamma_{it}$ depends on the $\mu$ parameter values for its past contracts. As the training procedure of the CANN model is iterative, the estimated $\mu$ values change at each iteration. Hence, it is crucial to update $\Sigma_{it}^{(\mu)}$ for each contract $(i, t)$ at every iteration. This updating procedure is carried out in step 2 of Algorithm~\ref{alg:gradient_descent_mvnb}.
\begin{algorithm}[ht]
    \SetAlgoLined
    \KwIn{Training dataset $\{(\boldsymbol{x}_{it}, y_{it})\}_{(i, t) \in \mathcal{T}_r}$, learning rate $\eta$, number of epochs $E$}
    \KwOut{Trained model parameters $\hat{\boldsymbol{\theta}}$, $\hat{\boldsymbol{\beta}}$ and $\hat{w}_\phi$}
    \BlankLine
    Initialize model parameters $\hat{\boldsymbol{\theta}} = \boldsymbol{0}$, $\hat{\boldsymbol{\beta}} = \hat{\boldsymbol{\beta}}^\text{MLE}$ and $\hat{w}_\phi$.\\
    Compute the number of past claims for each contract $(i, t)$: $\Sigma_{it}^{(y)} = \sum_{t' = 1}^{t-1} y_{it'}$.\\
    \BlankLine
    \For{epoch $\leftarrow$ 1 \KwTo $E$}{
        \begin{enumerate}
            \item For each contract $(i, t)$, apply the CANN regression function with current network parameters to compute the current estimated mean parameter $\hat{\mu}_{it}$:
            \begin{align*}
                \hat{\mu}_{it} = \mu^\text{CANN}(\boldsymbol{x}_{it}; \hat{\boldsymbol{\beta}}, \hat{\boldsymbol{\theta}}).
            \end{align*}
            \item Initialize or update the sum of past $\mu$ values for each contract $(i, t)$: $\hat{\Sigma}_{it}^{(\mu)} = \sum_{t' = 1}^{t-1} \hat{\mu}_{it}$.
            \item Compute the current estimated parameter $\hat{\phi}$:
            \begin{align*}
                \hat{\phi} = \zeta(\hat{w}_\phi).
            \end{align*}
            \item Compute the current estimated parameter $\hat{\alpha}_{it}$ and $\hat{\gamma}_{it}$:
            \begin{align*}
                \hat{\alpha}_{it} = \hat{\phi} + \Sigma_{it}^{(y)},\\
                \hat{\gamma}_{it} = \hat{\phi} + \hat{\Sigma}_{it}^{(\mu)}.
            \end{align*}
            \item Compute the empirical risk over the training dataset: 
            \begin{align*}
                \mathcal{R} = -\frac{1}{|\mathcal{T}_r|}\sum_{(i,t) \in \mathcal{T}_r} \ln \left[\frac{\Gamma(y_{it} + \hat{\alpha}_{it})}{y_{it}!\Gamma(\hat{\alpha}_{it})}\right] + \hat{\alpha}_{it} \ln\left[\frac{\hat{\gamma}_{it}}{\hat{\gamma}_{it} + \hat{\mu}_{it}}\right] + y_{it} \ln\left[\frac{\hat{\mu}_{it}}{\hat{\mu}_{it} + \hat{\gamma}_{it}}\right].
            \end{align*}
            \item Perform backpropagation to compute the gradients of $\mathcal{R}$ with respect to the network parameters: 
            \begin{align*}
                \nabla_{\boldsymbol{\beta}} \mathcal{R} \text{,}\quad \nabla_{\boldsymbol{\theta}} \mathcal{R} \quad\text{and}\quad \nabla_{w_\phi} \mathcal{R}.
            \end{align*}
            \item Perform gradient descent using the learning rate:
            \begin{align*}
                \hat{\boldsymbol{\beta}} \leftarrow \hat{\boldsymbol{\beta}} - \eta \nabla_{\hat{\boldsymbol{\beta}}} \mathcal{R},\\
                \hat{\boldsymbol{\theta}} \leftarrow \hat{\boldsymbol{\theta}} - \eta \nabla_{\hat{\boldsymbol{\theta}}} \mathcal{R},\\
                \hat{w}_\phi \leftarrow \hat{w}_\phi - \eta \nabla_{\hat{w}_\phi} \mathcal{R}.
            \end{align*}
        \end{enumerate}
    }
    \caption{Parameter estimation procedure -- MVNB CANN model}
    \label{alg:gradient_descent_mvnb}
\end{algorithm}

\section{Pratical Application with Telematics Data}\label{sec:application}

In this section, we explain how our CANN regression models are applied to our datastet. Additionally, we describe the application of the log-linear models, which serve as benchmark models in our analysis.

\subsection{Log-linear models}

The Poisson, negative binomial, and MVNB log-linear models are benchmarks for the Poisson, negative binomial, and MVNB CANN models. These models incorporate all 11 traditional risk factors from Table~\ref{tab:classic}, including the real distance driven (although not strictly classified as a traditional risk factor). For each contract $(i, t)$, these traditional risk factors are denoted by the vector $\boldsymbol{x}_{it}^{(\text{trad})}$. Notice that among the 11 traditional risk factors, 4 are categorical: \texttt{gender}, \texttt{marital\_status}, \texttt{pmt\_plan}, and \texttt{veh\_use}. For these risk factors, the approach involves initially grouping all rare categories, defined as those representing 5\% or less of the total number of observations, and labeling them as ``others.'' We then encode them numerically using dummy encoding. All the resulting traditional covariates are then centered and scaled. Moreover, \texttt{commute\_distance} contains missing values, which we fill in using median imputation.

Unlike neural networks, log-linear models do not have the ability to learn features directly from raw data. As a result, we must manually engineer features from the telematics data used by these models. These 13 telematics features, described in Table~\ref{tab:hc_telematics}, were specifically engineered from the telematics dataset as risk factors potentially correlated with the claiming risk. For each contract $(i, t)$, these numerical handcrafted telematics features are denoted by the vector $\boldsymbol{x}_{it}^{(\text{hand})}$. Note that these handcrafted telematics features are also centered and scaled prior to being input into the log-linear models. The regression function for the $\mu$ parameter can thus be written as
\begin{align}
    \mu\left(\boldsymbol{x}^{(\text{trad, hand})};\boldsymbol{\beta}\right) = \exp\left(\langle \boldsymbol{x}^{(\text{trad, hand})}, \boldsymbol{\beta} \rangle\right),
\end{align}
where $\boldsymbol{x}^{(\text{trad, hand})}$ is the concatenation of $\boldsymbol{x}^{(\text{trad})}$ and $\boldsymbol{x}^{(\text{hand})}$.
\begin{table}[ht]
    \centering
        \begin{tabular}{l l}
            \toprule 
            \textbf{Feature name} & \textbf{Description}\\ 
            \midrule
            \texttt{avg\_daily\_nb\_trips} & Average daily number of trips\\
            \texttt{frac\_expo\_evening} & Fraction of evening driving\tablefootnote{20h-0h}\\
            \texttt{frac\_expo\_fri\_sat} & Fraction of driving on Friday and Saturday\\
            \texttt{frac\_expo\_mon\_to\_thu} & Fraction of driving on Monday to Thursday\\
            \texttt{frac\_expo\_night} & Fraction of night driving\tablefootnote{0h-6h}\\ 
            \texttt{frac\_expo\_noon} & Fraction of midday driving\tablefootnote{11h-14h}\\ 
            \texttt{frac\_expo\_peak\_evening} & Fraction of evening rush hour driving\tablefootnote{17h-20h Monday to Friday}\\ 
            \texttt{frac\_expo\_peak\_morning} & Fraction of morning rush hour driving\tablefootnote{7h-9h Monday to Friday}\\
            \texttt{max\_trip\_max\_speed} & Maximum of the maximum speed of the trips\\ 
            \texttt{med\_trip\_avg\_speed} & Median of the average speeds of the trips\\ 
            \texttt{med\_trip\_distance} & Median of the distances of the trips\\
            \texttt{med\_trip\_max\_speed} & Median of the maximum speeds of the trips\\ 
            \texttt{prop\_long\_trip} & Proportion of long trips ($>100\text{km}$)\\
            \bottomrule 
        \end{tabular}
    \caption{Handcrafted telematics features extracted from the telematics dataset.} 
    \label{tab:hc_telematics} 
\end{table}
The parameters estimated on the training set are shown in Table~\ref{tab:coefs_art_3}. Notably, when using telematics information, the estimated $\phi$ parameter in the MVNB log-linear model is higher. A higher $\phi$ value brings the correcting factor in Equation~\ref{eq:esp_mvnb} closer to one, indicating reduced importance on past experience when telematics features are used. This underscores the relevance of the engineered telematics features.
\begin{table}[ht]
    \centering
    \begin{tabular}{l | *{3}{S[table-format=-1.4]} | *{3}{S[table-format=-1.4]}}
        \toprule 
        & \multicolumn{3}{c|}{\textbf{No telematics}} & \multicolumn{3}{c}{\textbf{With telematics}} \\
        \textbf{Parameters} & {Poisson} & {Negative binomial} & {MVNB} & {Poisson} & {Negative binomial} & {MVNB}\\ 
        \midrule
        \texttt{Intercept} & -2.8310 & -2.8311 & -2.8281 & -2.8454 & -2.8456 & -2.8430\\
        \midrule
        \texttt{annual\_distance} & 0.0273 & 0.0280 & 0.0288 & 0.0353 & 0.0359 & 0.0367\\
        \texttt{commute\_distance} & 0.0055 & 0.0055 & 0.0056 & 0.0159 &  0.0159 & 0.0157 \\
        \texttt{conv\_count\_3\_yrs\_minor} & 0.0470 & 0.0474 & 0.0469 & 0.0381 & 0.0384 & 0.0380\\
        \texttt{distance} & 0.1697 & 0.1706 & 0.1681 & 0.1244 & 0.1252 & 0.1232\\
        \texttt{expo} & 0.1945 & 0.1943 & 0.1957 & 0.1812 & 0.1813 & 0.1834\\
        \texttt{gender\_Male} & -0.0234 & -0.0238 & -0.0236 & -0.0409 & -0.0415 & -0.0415\\
        \texttt{marital\_status\_Single} & 0.0241 & 0.0243 & 0.0243 & 0.0194 & 0.0194 & 0.0192\\
        \texttt{marital\_status\_other} & 0.0342 & 0.0341 & 0.0341 &  0.0299 & 0.0297 & 0.0298\\
        \texttt{pmt\_plan\_EFT.Monthly} & 0.0963 & 0.0965 & 0.0969 & 0.0828 & 0.0830 & 0.0833\\
        \texttt{pmt\_plan\_Monthly} & 0.0856 & 0.0854 & 0.0850 & 0.0773 & 0.0771 & 0.0768\\
        \texttt{pmt\_plan\_other} & 0.0134 & 0.0135 & 0.0131 & 0.0111 & 0.0111 & 0.0107\\
        \texttt{veh\_age} & -0.1552 & -0.1543 & -0.1540 & -0.1433 & -0.1425 & -0.1422\\
        \texttt{veh\_use\_other} & -0.0085 & -0.0083 & -0.0084 & -0.0100 & -0.0098 & -0.0100\\
        \texttt{veh\_use\_pleasure} & -0.0025 & -0.0023 & -0.0027 & -0.0014 & -0.0013 &  -0.0018\\
        \texttt{years\_licensed} & -0.1061 & -0.1064 & -0.1076 & -0.0538 & -0.0539 & -0.0547\\
        \midrule
        \texttt{avg\_daily\_nb\_trips} & {--} & {--} & {--} &  0.0428 & 0.0424 & 0.0411\\
        \texttt{frac\_expo\_evening} & {--} & {--} & {--} & 0.0734 & 0.0738 & 0.0741\\
        \texttt{frac\_expo\_fri\_sat} & {--} & {--} & {--} & 0.0290 & 0.0288 & 0.0294\\
        \texttt{frac\_expo\_mon\_to\_thu} & {--} & {--} & {--} & 0.0854 & 0.0852 & 0.0857\\
        \texttt{frac\_expo\_night} & {--} & {--} & {--} & 0.0192 & 0.0193 & 0.0198\\
        \texttt{frac\_expo\_noon} & {--} & {--} & {--} & 0.0103 & 0.0100 & 0.0092\\
        \texttt{frac\_expo\_peak\_evening} & {--} & {--} & {--} & 0.0049 & 0.0047 & 0.0046\\
        \texttt{frac\_expo\_peak\_morning} & {--} & {--} & {--} & 0.0072 & 0.0073 & 0.0071\\
        \texttt{max\_trip\_max\_speed} & {--} & {--} & {--} & 0.1084 & 0.1087 & 0.1079\\
        \texttt{med\_trip\_avg\_speed} & {--} & {--} & {--} & -0.1465 & -0.1470 & -0.1472\\
        \texttt{med\_trip\_distance} & {--} & {--} & {--} & 0.0082 & 0.0088 & 0.0081\\
        \texttt{med\_trip\_max\_speed} & {--} & {--} & {--} & 0.0725 & 0.0723 & 0.0736\\
        \texttt{prop\_long\_trip} & {--} & {--} & {--} & 0.0310 & 0.0314 & 0.0322\\
        \midrule
         $\phi$ & {--} & 2.8397 & 3.4868 & {--} & 3.1193 & 3.9119\\
        \bottomrule 
    \end{tabular}
    \caption{Estimated parameters of the log-linear models on the training set.} 
    \label{tab:coefs_art_3} 
\end{table}

\subsection{CANN models}

For the CANN regression models, we extract low-level descriptor vectors that are specifically designed to accurately describe the driving patterns within a particular contract, at least with the dataset we have. We expect the MLP component within the CANN models to learn meaningful high-level features from these low-level vectors. The hope is that the learned features in the hidden layers will be more relevant than the handcrafted features of Table~\ref{tab:hc_telematics}. Each contract $(i, t)$ is described by the following descriptor vectors, which provide a summary of its telematics information:
\begin{align*}
    \boldsymbol{x}_{it}^{(h)} &= \left(x_{it, 1}^{(h)}, \dots, x_{it, 24}^{(h)}\right) \in \mathbb{R}^{24},\\
    \boldsymbol{x}_{it}^{(d)} &= \left(x_{it, 1}^{(d)}, \dots, x_{it, 7}^{(d)}\right) \in \mathbb{R}^{7},\\
    \boldsymbol{x}_{it}^{(a)} &= \left(x_{it, 1}^{(a)}, \dots, x_{it, 14}^{(a)}\right) \in \mathbb{R}^{14},\\
    \boldsymbol{x}_{it}^{(m)} &= \left(x_{it, 1}^{(m)}, \dots, x_{it, 16}^{(m)}\right) \in \mathbb{R}^{16},\\
    \boldsymbol{x}_{it}^{(k)} &= \left(x_{it, 1}^{(k)}, \dots, x_{it, 10}^{(k)}\right) \in \mathbb{R}^{10}.\\
\end{align*}
\begin{itemize}
    \item The elements in vector $\boldsymbol{x}_{it}^{(h)}$ represent the fraction of driving during each of the 24 hours of the day. Therefore, $x_{it, j}^{(h)}$ is the fraction of driving during the $j^\text{th}$ hour of the day for contract $(i, t)$. 
    \item The elements in vector $\boldsymbol{x}_{it}^{(d)}$ represent the fraction of driving during each of the 7 days of the week. Therefore, $x_{it, j}^{(d)}$ is the fraction of driving during the $j^\text{th}$ day of the week for contract $(i, t)$. Monday, Tuesday, Wednesday, Thursday, Friday, Saturday, and Sunday are denoted by $j = 1, 2, 3, 4, 5, 6, 7$, respectively.
    \item The elements in vector $\boldsymbol{x}_{it}^{(a)}$ represent the fraction of trips made in different average speed slots. For instance, $x_{it, j}^{(a)}$ denotes the fraction of trips made at an average speed between $10(j-1)$ and $10j$ kilometers per hour.
    \item The elements in vector $\boldsymbol{x}_{it}^{(m)}$ represent the fraction of trips made in different maximum speed slots. For instance, $x_{it, j}^{(m)}$ denotes the fraction of trips made where the maximum speed reached falls between $10(j-1)$ and $10j$ kilometers per hour.
    \item The elements in vector $\boldsymbol{x}_{it}^{(k)}$ represent the fraction of trips made in different distance slots. For instance, $x_{it, j}^{(k)}$ denotes the fraction of trips between $5(j-1)$ and $5j$ kilometers.
\end{itemize}
These descriptor vectors capture specific aspects of the driving patterns, such as hourly, weekly, average speed, and maximum speed distribution, providing valuable information for the MLPs. Since MLPs can only accept vectors as input, we concatenate these four vectors into a global telematics vector:
\begin{align*}
    \boldsymbol{x}_{it}^{(tele)} = \left(\boldsymbol{x}_{it}^{(h)}, \boldsymbol{x}_{it}^{(d)}, \boldsymbol{x}_{it}^{(a)}, \boldsymbol{x}_{it}^{(m)}, \boldsymbol{x}_{it}^{(k)}\right).
\end{align*}
We incorporate this telematics vector into the MLP component of the CANN models, together with the traditional risk factors $\boldsymbol{x}^{(trad)}$, enabling interactions between telematics and traditional inputs. In contrast, the log-linear part of the CANN models only includes the traditional risk factors due to the difficulty of processing low-level information. The regression function for the $\mu$ parameter can thus be written as
\begin{align}
\mu^\text{CANN}\left(\boldsymbol{x}^{(\text{trad, tele})}; \boldsymbol{\beta}, \boldsymbol{\theta}\right) &= \zeta\left\{\langle\boldsymbol{x}^\text{trad}, \boldsymbol{\beta}\rangle + \boldsymbol{a}^{(L - 1)}(\boldsymbol{x}^\text{trad,tele}; \boldsymbol{\theta})\right\}.
\end{align}
where $\boldsymbol{x}^{(\text{trad, tele})}$ is the concatenation of $\boldsymbol{x}^{(\text{trad})}$ and $\boldsymbol{x}^{(\text{tele})}$.

The CANN models are trained using the \texttt{torch} library in the \texttt{R} programming language, using mini-batch gradient descent with 256 observations per batch. The optimizer we use to perform gradient descent is the Adam optimizer, which is a fairly popular choice for training neural networks. Additionally, we use the reduce-on-plateau learning rate scheduler, which dynamically adjusts the learning rate based on the model's performance, automatically reducing it when the improvement plateaus, allowing for better optimization and convergence during training. For the MLP component of our CANN models, we opt for 3 hidden layers $(L = 5)$ with 128, 64, and 32 hidden units, respectively $(n_1 = 128, n_2 = 64, n_3 = 32)$. We choose the rectified linear unit (ReLU) as the activation function $\phi(\cdot)$ used in the hidden layers. Additionally, we add batch normalization and dropout layers in-between fully connected layers. Batch normalization applies a normalization transformation to the input of a layer by subtracting the mini-batch mean and dividing by the mini-batch standard deviation. By maintaining a stable mean and variance throughout the network, it can mitigate the vanishing or exploding gradients problem, enabling more effective and efficient training. The dropout layers, on the other hand, serve as a regularization technique that helps prevent overfitting. During training, dropout randomly sets a fraction of the hidden units of a given hidden layer to zero at each iteration, which forces the network to learn redundant representations and reduces the reliance on specific features. This regularization technique improves the model's ability to generalize well to unseen data.

\subsection{CANN hyperparameter tuning}

To maximize the performance of our CANN models, we use grid search for hyperparameter tuning, with the average loss observed on the validation dataset $\mathcal{V}_a$ as our optimization criterion. Additionally, we incorporate a regularization technique known as ``early stopping'' to determine the best number of epochs. This approach allows us to prevent overfitting and select the optimal number of epochs based on the lowest average loss achieved during training. We focus on three key hyperparameters: \texttt{p}, which represents the probability of dropout in the dropout layers, \texttt{l\_start}, denoting the initial learning rate used in the reduce-on-plateau learning rate scheduler, and \texttt{factor}, indicating the factor by which the learning rate is multiplied upon reaching a plateau. A plateau is the point where there is no observed improvement in the validation loss for two consecutive epochs. We compute the average validation loss for all 45 combinations derived from the following hyperparameter values:
\begin{itemize}
    \item \texttt{l\_start}: $0.00001, 0.00005, 0.0001, 0.0005, 0.001$
    \item \texttt{factor}: $0.3, 0.4, 0.5$
    \item \texttt{p}: $0.2, 0.3, 0.4$.
\end{itemize}
Remember that the network parameters in the classical components of the CANN models are initialized with the maximum likelihood estimators of the corresponding log-linear model, which is why we use relatively small learning rates. At the initialization stage, the network already produces reasonable predictions, reducing the need for large gradient descent steps. The validation loss for each of the 45 combinations and the three specifications is presented in Table~\ref{tab:coarse_ht}. It is worth noting that each model is trained for 30 epochs, and as early stopping is employed, the displayed average validation loss is based on the optimal number of epochs, which can be less than 30.
\small
\begin{table}[ht]
    \centering
    \begin{tabular}{ccc|ccc|ccc}
        \toprule 
        \multicolumn{3}{c|}{\textbf{Hyperparameter values}} & \multicolumn{3}{c|}{\textbf{Average validation loss}} & \multicolumn{3}{c}{\textbf{Number of epochs}}\\
        \midrule
        \texttt{l\_start} & \texttt{factor} & \texttt{p} & Poisson & Negative binomial & MVNB & Poisson & Negative binomial & MVNB \\
        \midrule        
        0.00001 & 0.3 & 0.2 & 0.2357 & 0.2351 & 0.2350 & 8 & 16 & 17 \\
        0.00001 & 0.3 & 0.3 & 0.2354 & 0.2350 & 0.2350 & 10 & 24 & 22 \\
        0.00001 & 0.3 & 0.4 & 0.2353 & 0.2351 & 0.2349 & 18 & 29 & 30 \\
        0.00001 & 0.4 & 0.2 & 0.2358 & 0.2351 & 0.2350 & 8 & 16 & 17 \\
        0.00001 & 0.4 & 0.3 & 0.2355 & 0.2350 & 0.2350 & 10 & 24 & 22 \\
        0.00001 & 0.4 & 0.4 & 0.2353 & 0.2351 & 0.2349 & 18 & 29 & 30\\
        0.00001 & 0.5 & 0.2 & 0.2360 & 0.2351 & 0.2350 & 8 & 16 & 17 \\
        0.00001 & 0.5 & 0.3 & 0.2356 & 0.2350 & 0.2350 & 10 & 24 & 22 \\
        0.00001 & 0.5 & 0.4 & 0.2354 & 0.2351 & 0.2349 & 18 & 29 & 30 \\
        \midrule
        0.00005 & 0.3 & 0.2 & 0.2355 & 0.2351 & 0.2349 & 2 & 4 & 4 \\
        0.00005 & 0.3 & 0.3 & 0.2355 & 0.2351 & 0.2349 & 3 & 5 & 5 \\
        0.00005 & 0.3 & 0.4 & 0.2363 & 0.2352 & 0.2351 & 4 & 8 & 7 \\
        0.00005 & 0.4 & 0.2 & 0.2355 & 0.2351 & 0.2349 & 2 & 4 & 4 \\
        0.00005 & 0.4 & 0.3 & 0.2355 & 0.2351 & 0.2349 & 3 & 5 & 5 \\
        0.00005 & 0.4 & 0.4 & 0.2365 & 0.2352 & 0.2351 & 4 & 8 & 7 \\
        0.00005 & 0.5 & 0.2 & 0.2355 & 0.2351 & 0.2349 & 2 & 4 & 4 \\
        0.00005 & 0.5 & 0.3 & 0.2355 & 0.2351 & 0.2349 & 3 & 5 & 5 \\
        0.00005 & 0.5 & 0.4 & 0.2369 & 0.2352 & 0.2351 & 4 & 8 & 7 \\ 
        \midrule
        0.0001 & 0.3 & 0.2 & 0.2354 & 0.2349 & 0.2349 & 1 & 3 & 3 \\
        0.0001 & 0.3 & 0.3 & 0.2354 & 0.2350 & 0.2348 & 2 & 3 & 3 \\
        0.0001 & 0.3 & 0.4 & 0.2371 & 0.2352 & 0.2350 & 2 & 5 & 4 \\
        0.0001 & 0.4 & 0.2 & 0.2354 & 0.2349 & 0.2349 & 1 & 3 & 3 \\
        0.0001 & 0.4 & 0.3 & 0.2354 & 0.2350 & 0.2348 & 2 & 3 & 3 \\
        0.0001 & 0.4 & 0.4 & 0.2374 & 0.2352 & 0.2350 & 2 & 5 & 4 \\
        0.0001 & 0.5 & 0.2 & 0.2354 & 0.2349 & 0.2349 & 1 & 3 & 3 \\
        0.0001 & 0.5 & 0.3 & 0.2354 & 0.2350 & 0.2348 & 2 & 3 & 3 \\
        0.0001 & 0.5 & 0.4 & 0.2378 & 0.2352 & 0.2350 & 2 & 5 & 4 \\        
        \midrule
        0.0005 & 0.3 & 0.2 & 0.2356 & 0.2350 & 0.2350 & 1 & 1 & 1 \\
        0.0005 & 0.3 & 0.3 & 0.2354 & 0.2352 & 0.2350 & 2 & 1 & 1 \\
        0.0005 & 0.3 & 0.4 & 0.2358 & 0.2352 & 0.2350 & 2 & 4 & 3 \\
        0.0005 & 0.4 & 0.2 & 0.2356 & 0.2350 & 0.2350 & 1 & 1 & 1 \\
        0.0005 & 0.4 & 0.3 & 0.2354 & 0.2352 & 0.2350 & 2 & 1 & 1 \\
        0.0005 & 0.4 & 0.4 & 0.2358 & 0.2352 & 0.2350 & 2 & 4 & 3 \\
        0.0005 & 0.5 & 0.2 & 0.2356 & 0.2350 & 0.2350 & 1 & 1 & 1 \\
        0.0005 & 0.5 & 0.3 & 0.2354 & 0.2352 & 0.2350 & 2 & 1 & 1 \\
        0.0005 & 0.5 & 0.4 & 0.2358 & 0.2352 & 0.2350 & 2 & 4 & 3 \\
        \midrule
        0.001 & 0.3 & 0.2 & 0.2358 & 0.2353 & 0.2351 & 1 & 1 & 1 \\
        0.001 & 0.3 & 0.3 & 0.2362 & 0.2352 & 0.2349 & 1 & 1 & 1 \\
        0.001 & 0.3 & 0.4 & 0.2362 & 0.2350 & 0.2349 & 2 & 2 & 2 \\
        0.001 & 0.4 & 0.2 & 0.2358 & 0.2353 & 0.2351 & 1 & 1 & 1 \\
        0.001 & 0.4 & 0.3 & 0.2362 & 0.2352 & 0.2349 & 1 & 1 & 1 \\
        0.001 & 0.4 & 0.4 & 0.2362 & 0.2350 & 0.2349 & 2 & 2 & 2 \\
        0.001 & 0.5 & 0.2 & 0.2358 & 0.2353 & 0.2351 & 1 & 1 & 1 \\
        0.001 & 0.5 & 0.3 & 0.2362 & 0.2352 & 0.2349 & 1 & 1 & 1 \\
        0.001 & 0.5 & 0.4 & 0.2362 & 0.2350 & 0.2349 & 2 & 2 & 2 \\
        \bottomrule 
    \end{tabular}
    \caption{Coarse hyperparameter tuning for the CANN models. The training process is stopped after 30 epochs. The provided validation loss corresponds to the optimal number of epochs, consistent with the early stopping procedure.}
    \label{tab:coarse_ht}
\end{table}
\normalsize
As can be seen, for all learning rates higher than 0.00001, the minimum average validation loss is achieved after a very small number of epochs, indicating that the network learns too quickly. Although the negative binomial and MVNB models perform best at a learning rate of 0.0001, we believe that with more epochs, we could achieve a lower average loss with a learning rate of 0.00001. This is particularly true since the average losses are quite similar for \texttt{lr\_start} = 0.00001 and \texttt{lr\_start} = 0.0001. When examining the first 9 rows of Table~\ref{tab:coarse_ht}, it becomes apparent that the \texttt{factor} hyperparameter has a negligible effect on the validation loss. On the other hand, the \texttt{p} hyperparameter only seems to have an impact on the validation loss for the Poisson model, performing best when \texttt{p} = 0.4. Although the dropout rate does not significantly affect the performance for both the negative binomial and MVNB models, we also choose \texttt{p} = 0.4 for these two models since the best performance is achieved at a high number of epochs (29 and 30 epochs, respectively). This suggests that with more epochs, there is potential for further performance improvement. Therefore, we select \texttt{lr\_start} = 0.00001, \texttt{factor} = 0.3, and \texttt{p} = 0.4 as the hyperparameters for all three specifications. We train the models again on the training set, this time for 100 epochs. The performance of the three models on the validation set is displayed in Table~\ref{tab:res_valid}.
\begin{table}[ht]
    \centering
    \begin{tabular}{l|cc}
        \toprule
        \textbf{Specification} & \textbf{Average validation loss} & \textbf{Number of epochs}\\
        \midrule
        Poisson & 0.2352 & 35 \\
        Negative binomial & 0.2351 & 35 \\
        MVNB & 0.2349 & 35 \\
        \bottomrule
    \end{tabular}
    \caption{Optimal CANN models' performance on the validation set.}
    \label{tab:res_valid}
\end{table}
As can be seen, all 3 specifications require 35 epochs to minimize the average validation loss. 

\section{Analyzes}\label{sec:analyzes}

\subsection{Performance assessment on the testing set}

After carefully tuning the hyperparameters of our CANN models, we have at hand promising claim count models that are now nearing implementation. The next crucial step is to estimate their generalization capabilities accurately. To achieve this, we cannot rely on the validation set, as it has been extensively used during the hyperparameter tuning process. Instead, we assess the models' generalization performance using the testing set $\mathcal{T}_e$, which has remained untouched until now. Using this independent dataset, we can estimate the models’ predictive performance on unseen data points and determine their suitability for real-world applications. Furthermore, we perform a comparative analysis between the CANN and the benchmark models, namely the log-linear models that use telematics information in the form of handcrafted telematics features. This comparative assessment allows us to evaluate our CANN models' relative performance and effectiveness against established approaches. In order to fully capture the value of telematics data, we also evaluate the performance of all 6 models (Poisson, negative binomial and MVNB log-linear and CANN models) using only the 11 traditional risk factors as covariates. In the CANN models, the MLP component therefore only comprises the 11 traditional risk factors. This analysis helps us understand the contribution of telematics information in improving the predictive power of the models.

All 12 models are trained on the learning set, and their performance is evaluated on the testing set. To assess the performance, we employ 3 different scoring rules, namely the Poisson deviance, the logarithmic score, and the squared error. For each scoring rule, we compute the average value on the testing set. To assess the magnitude of the achieved performance, we begin by calculating the average scoring rule values for a baseline model. This baseline model is defined as a homogeneous Poisson log-linear model, where the estimation of the mean (and variance) parameter $\mu$ is estimated by the average number of claims per contract observed in the learning set:
\begin{align}
    \hat{\mu} = \frac{1}{|\{\mathcal{T}_r, \mathcal{V}_a\}|}\sum_{(i, t)\in \{\mathcal{T}_r, \mathcal{V}_a\}} y_{it}.
\end{align}
The average scoring rule values for this baseline model on the testing set are reported in Table~\ref{tab:res_baseline}.
\begin{table}[ht]
    \centering
    \begin{tabular}{l|c}
        \toprule
        \textbf{Scoring rule} & \textbf{Baseline model} \\
        \midrule
        Poisson deviance & 0.3682 \\
        Logarithmic score & 0.2470 \\
        Squared error & 0.0697 \\
        \bottomrule
    \end{tabular}
    \caption{Performance of the baseline model on the testing set.}
    \label{tab:res_baseline}
\end{table}
We can then evaluate the performance of each of the 6 models in terms of percentage improvement over the baseline model, as shown in Table~\ref{tab:test_results}.
\begin{table}[ht]
    \centering
    \begin{tabular}{l|cc|cc}
        \toprule
         & \multicolumn{2}{c|}{\textbf{No telematics}} & \multicolumn{2}{c}{\textbf{With telematics}}\\
        \textbf{Scoring rule} & Log-linear model & CANN model & Log-linear model & CANN model \\
        \midrule
         & \multicolumn{4}{c}{Poisson} \\
        \midrule
        Poisson deviance & 5.23 \% & 5.53 \% & 5.68 \% & 5.78 \% \\
        Logarithmic score & 3.90 \% & 4.12 \% & 4.23 \% & 4.31 \% \\
        Squared error & 2.10 \% & 2.26 \% & 2.30 \% & 2.38 \% \\
        \midrule
         & \multicolumn{4}{c}{Negative binomial} \\
        \midrule
        Poisson deviance & 5.24 \% & 5.58 \% & 5.68 \% & 5.81 \% \\
        Logarithmic score & 3.99 \% & 4.24 \% & 4.31 \% & 4.41 \% \\
        Squared error & 2.10 \% & 2.27 \% & 2.30 \% & 2.37 \% \\
        \midrule
         & \multicolumn{4}{c}{MVNB} \\
        \midrule
        Poisson deviance & 5.36 \% & 5.65 \% & 5.79 \% & 5.90 \% \\
        Logarithmic score & 4.07 \% & 4.27 \% & 4.38 \% & 4.46 \% \\
        Squared error & 2.13 \% & 2.29 \% & 2.34 \% & 2.41 \% \\
        \bottomrule
    \end{tabular}
    \caption{Performance comparison of the CANN models and their corresponding log-linear benchmark model on the testing set.}
    \label{tab:test_results}
\end{table}
As can be seen, our CANN models consistently outperform their corresponding log-linear benchmark models across all scoring rules. Moreover, our longitudinal MVNB CANN model offers a significative improvement over both Poisson and negative binomial distributions, suggesting a substantial dependence among contracts within a vehicle.

\subsection{Permutation feature importance and partial dependence plots}

One substantial drawback of neural networks is their difficulty of interpretation. However, researchers have developed tools to shed light on the inner workings of these black box algorithms. Two particularly useful tools in this context are permutation feature importance and partial dependence plots.

\begin{figure}[ht]
    \centering
    \includegraphics[width = \textwidth]{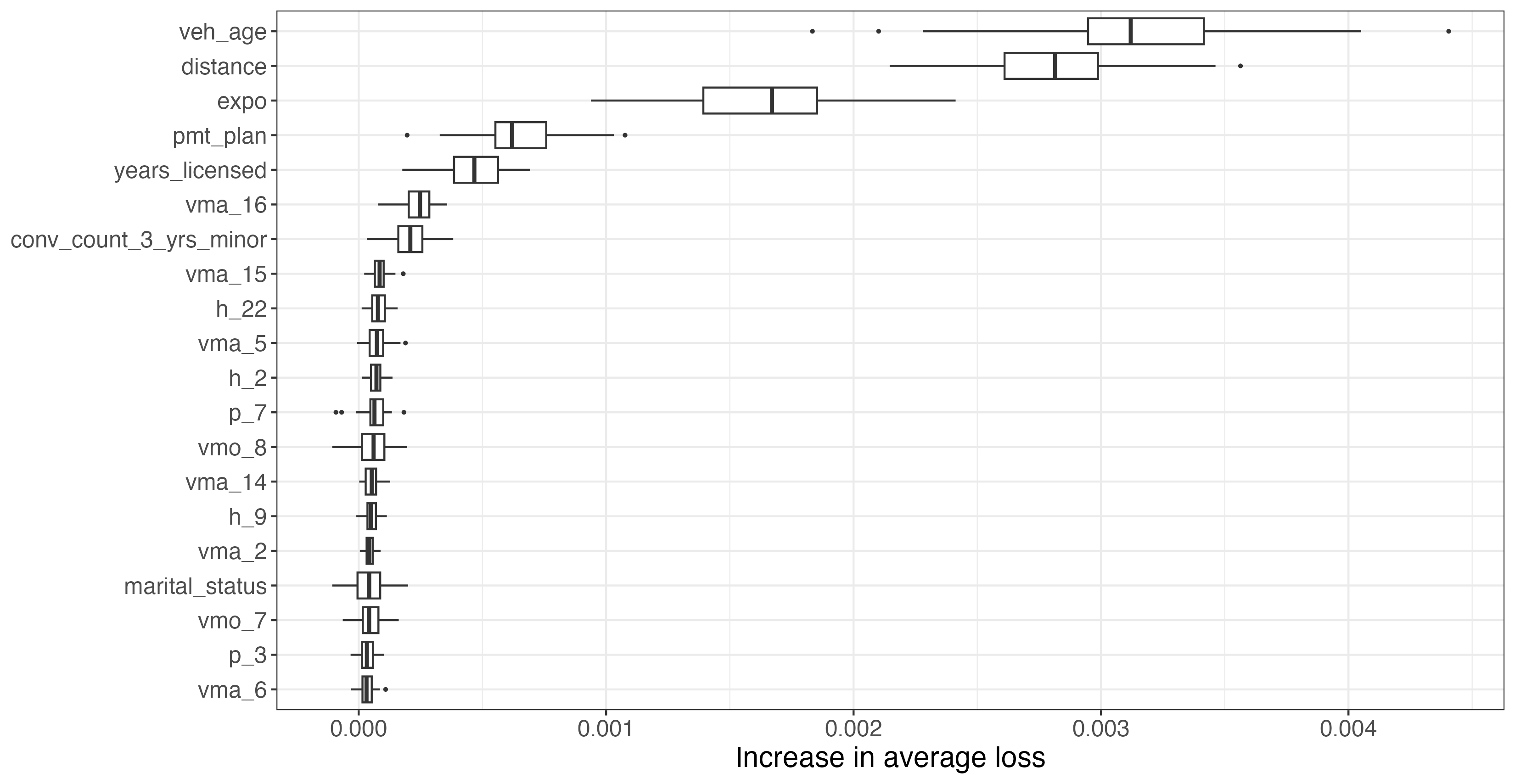}
    \caption{Importance scores of the 20 most important variables obtained for the MVNB CANN model.}
    \label{fig:imp}
\end{figure}

Permutation feature importance is a model-agnostic technique that computes an importance score for each input (or variable) in a supervised learning algorithm. It achieves this by randomly permuting the values of a specific input while holding the other inputs constant and observing the resulting effect on the model's performance. By comparing the original model's performance with the permuted performance, we can determine the variable's importance relative to the chosen performance metric. Suppose we have a trained model and a holdout sample for evaluation purposes. We can initially score the model on this sample and measure its performance using a chosen metric, such as the average loss. Let us denote the average loss obtained with the original holdout sample as $\ell_{\text{original}}$. To assess the importance of input $j$ in the prediction process, we randomly shuffle the values of input $j$ in the holdout sample and rescore the model. This process yields a new average loss, denoted as $\ell_{\text{permuted}}^{(j)}$, where the superscript $j$ indicates that input $j$ has been permuted. If input $j$ is indeed important for the model's prediction, the permuted average loss $\ell_{\text{permuted}}^{(j)}$ is expected to be greater than the original average loss $\ell_{\text{original}}$. This suggests that permuting the values of input $j$ has a detrimental effect on the model's performance. To obtain an importance score for input $j$, we can compute the difference between the permuted average loss and the original average loss, resulting in the feature importance score $\text{FI}_j$:
\begin{align*}
    \text{FI}_j = \ell_{permuted}^{(j)} - \ell_{original}.
\end{align*}
To obtain a more reliable estimate of the importance score, this procedure can be repeated a certain number of times for input $j$, creating a distribution of the increase or decrease in the average loss. The whole procedure can then be repeated for all inputs. In Figure~\ref{fig:imp}, the importance scores of the 20 most important variables for our best model, the MVNB CANN, are visualized using boxplots. Please note that the names used for the telematics inputs in Figure~\ref{fig:imp} differ from the introduced notation. However, a translation table is provided in Table~\ref{tab:translation} of Appendix~\ref{app:A_art_3} to clarify the correspondence between the names used and the introduced notation. Each boxplot represents the distribution of the 100 importance scores assigned to a specific input obtained by shuffling and assessing the model 100 times. The performance metric used is the average cross-entropy loss.
The analysis reveals interesting findings regarding the claim count model. As can be seen, the top 5 most important variables are from our set of 11 traditional risk factors. Notably, \texttt{veh\_age}, \texttt{distance}, and \texttt{expo} play a significant role in the model's performance. When it comes to telematics inputs, those related to maximum speed demonstrate a substantial impact on the model's performance. Particularly, \texttt{vma\_16}, representing the fraction of trips made at a maximum speed exceeding 150 kilometers per hour, stands out as the most important input. In general, the fraction of trips made at high maximum speeds, such as \texttt{vma\_14}, \texttt{vma\_15}, and \texttt{vma\_16}, proves to be valuable for predicting claims. Additionally, it is interesting to observe that \texttt{h\_22} and \texttt{h\_2}, which represent the fraction of driving during night hours, contribute substantially to the assessment of risk. Importantly, the \texttt{gender} variable, often used by insurers as a risk factor, is rendered useless in the presence of telematics inputs. It ranks as the $\text{70}^\text{th}$ most important variable (not showed in Figure~\ref{fig:imp}), indicating its insignificance in the model's predictive power.

\begin{figure}[ht]
    \centering
    \includegraphics[width = \textwidth]{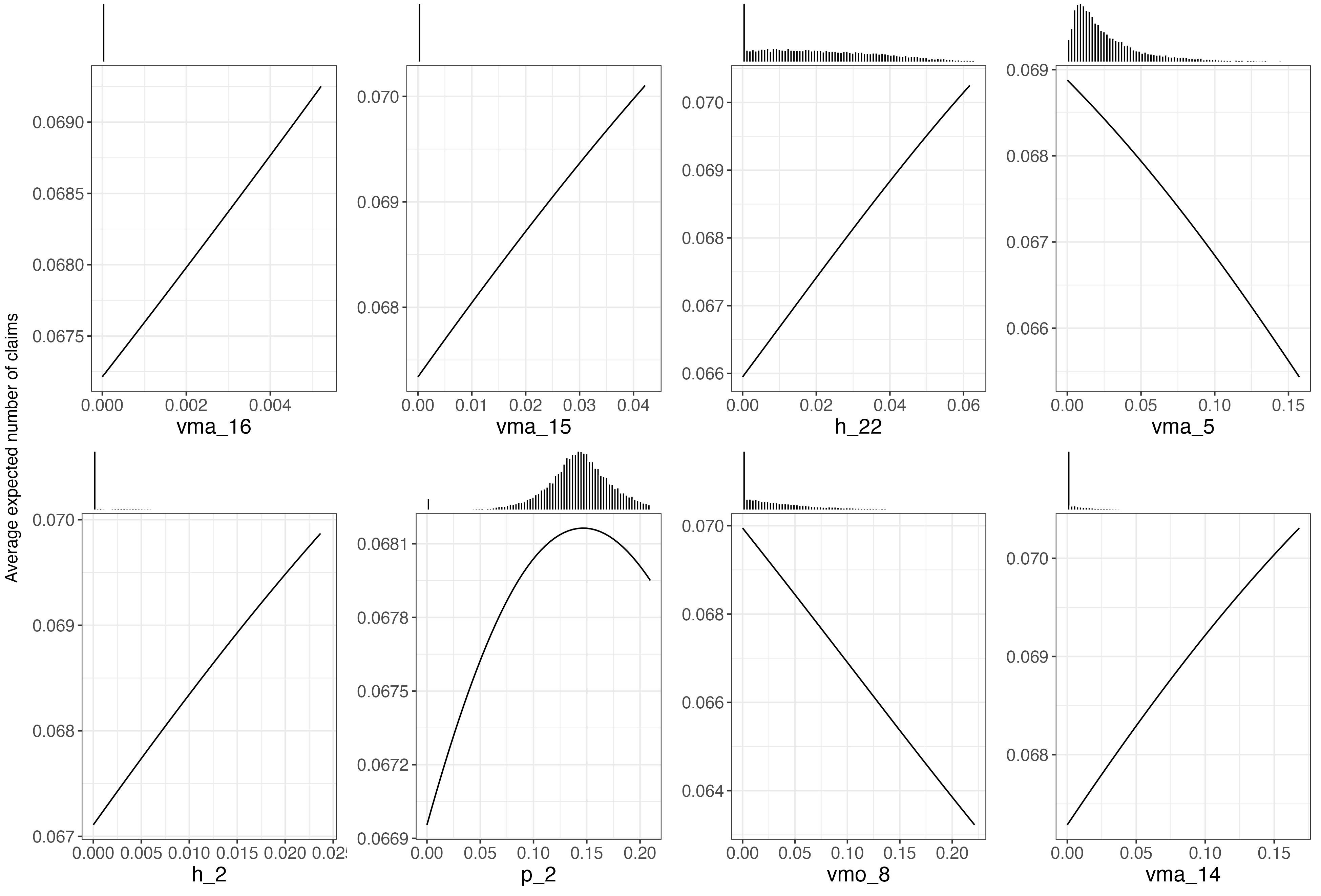}
    \caption{Partial dependence plots showcasing the 8 most important telematics inputs in the MVNB CANN model. The histogram above each line plots shows the input's distribution.}
    \label{fig:pdp}
\end{figure}

Partial Dependence Plots (PDP) are valuable tools for understanding the relationship between a specific input variable and the output of a supervised learning model. PDPs are also model-agnostic, meaning they can be applied to different types of models. They provide insights into how changes in a particular input variable influence the model's predictions while keeping all other variables at fixed values. In other words, they illustrate the marginal effect of an input variable on the predicted outcome. To compute a PDP for a specific input variable $j$, the process involves the following steps. First, a grid of values is defined to cover the entire or plausible range of the variable's values. Next, while holding all other variables fixed, the input vector in the holdout dataset is sequentially replaced with each value from the defined grid. Subsequently, predictions are obtained using the trained model on the modified holdout dataset for each value. By plotting the input variable values on the x-axis and the corresponding average prediction on the y-axis, the resulting PDP visually showcases the relationship between the input variable and the model's predictions. Figure~\ref{fig:pdp} displays the PDPs of the 8 most important telematics inputs in the MVNB CANN model. 
The plots reveal that the risk, expressed as the expected number of claims, appears to increase in a linear fashion with the proportion of trips made at high maximum speeds, indicated by the input variables \texttt{vma\_14}, \texttt{vma\_15}, and \texttt{vma\_16}. Additionally, there appears to be a positive linear association between the expected number of claims and the proportion of driving taking place during nighttime hours, specifically between 9 p.m. and 10 p.m. (h\_22) and between 1 a.m. and 2 a.m. (h\_2). It is important to emphasize that when interpreting partial dependence plots, caution must be exercised, as the procedure assumes that the input variables are independent of each other. In particular, the interpretation of the PDPs related to the fraction of driving on Tuesdays (\texttt{p\_2}) is challenging due to the correlation between the proportions of driving on different days of the week. For instance, if an insured individual drives in smaller proportions on Tuesdays, they will systematically drive in larger proportions on other days of the week.

\section{Conclusions}\label{sec:conclusions_article_3}

In this study, we developed three novel claim count regression models leveraging telematics data in the form of trip summaries. Our models are based on the Combined Actuarial Neural Network architecture, specifically designed to address actuarial problems and harness rich and complex information such as data provided by telematics technology. One key aspect of our work is the adaptation of the CANN architecture to accommodate the MVNB distribution specification. This adaptation allows us to effectively capture the time dependence between insurance contracts, which is important for accurately modeling claim counts. Furthermore, our findings highlight the importance of telematics inputs related to the maximum speed reached during trips in the claim count models. With partial dependence plots, we found that claim frequency is positively correlated with the fraction of trips made at high maximum speeds. Overall, the new approaches developed in this article represent a significant advancement in accurately modeling claim counts and enhancing the performance of predictive models in the context of usage-based insurance. Remarkably, the CANN regression models consistently outperform traditional log-linear models using handcrafted telematics features, as demonstrated by the superior performance across three performance metrics. These results are further supported by the use of a proper machine learning methodology that effectively prevents data leakage and mitigates the risk of producing falsely optimistic results.

While the available telematics data has been instrumental in improving our claim count models, we believe that further improvement can be achieved with access to richer data. For instance, if second-by-second data or additional information such as harsh acceleration/braking and distracted driving were accessible, we believe the performance could be further improved. Depending on the data format, different types of neural networks, such as convolutional and recurrent neural networks, could be used as the network component in the CANN models. Additionally, we acknowledge that with more time and computational power, a more comprehensive fine-tuning process of the CANN models could yield even better results than what we achieved. Notably, we were constrained in adjusting the number of hidden layers and units in the MLP components of the CANN models due to time and computational limitations. Moreover, a more advanced tuning method, beyond the grid search approach used in this study, could be employed to optimize model performance. In this study, we used the MVNB distribution as our longitudinal specification. However, alternative longitudinal specifications, such as the beta-binomial distribution, exist and could be easily implemented as they share similarities with the MVNB specification. Finally, it would be interesting to conduct further research investigating the impact of using a longitudinal model on telematics variables. It is expected that the importance of certain telematics variables would decrease when considering past claim history, as this historical data can provide insights into the claiming risk of an insured.


\clearpage

\section*{Acknowledgement}

The authors gratefully acknowledge The Co-operators for their generous financial support and for providing the data used in this paper through the Co-operators Chair in Actuarial Risk Analysis. Additionally, the authors would like to extend their sincere appreciation to Marc Morin from the Research and Innovation team at The Co-operators for his invaluable assistance with the \texttt{torch} library.

\section*{Funding}

The authors thank The Co-operators, the Natural Sciences and Engineering Research Council of Canada and Les fonds de recherche du Québec for funding.

\bibliographystyle{apalike}  
\bibliography{main}


\clearpage
\appendix

\section{Telematics Input Names Translation}\label{app:A_art_3}

\begin{table}[ht]
    \centering
    \begin{tabular}{ccl}
        \toprule
        \textbf{Introduced notation} & \textbf{Notation in the plots} & \textbf{Description}\\
        \midrule
        $x_{it, 1}^{(h)}$ & \texttt{h\_1} & Fraction of driving between midnight and 1 a.m.\\
        $x_{it, 2}^{(h)}$ & \texttt{h\_2}  & Fraction of driving between 1 a.m. and 2 a.m.\\
        \vdots & \vdots & \vdots \\
        $x_{it, 24}^{(h)}$ & \texttt{h\_24}  & Fraction of driving between 11 p.m. and midnight\\
        \cmidrule(l){1-3}\\
        $x_{it, 1}^{(d)}$ & \texttt{p\_1}  & Fraction of driving on Mondays\\
        $x_{it, 2}^{(d)}$ & \texttt{p\_2}  & Fraction of driving on Tuesdays\\
        \vdots & \vdots & \vdots \\
        $x_{it, 7}^{(d)}$ & \texttt{p\_7}  & Fraction of driving on Sundays\\
        \cmidrule(l){1-3} \\ 
        $x_{it, 1}^{(a)}$ & \texttt{vmo\_1}  & Fraction of trips with average speed between 0 and 10 kph\\
        $x_{it, 2}^{(a)}$ & \texttt{vmo\_2}  & Fraction of trips with average speed between 10 and 20 kph\\
        \vdots & \vdots & \vdots \\
        $x_{it, 14}^{(a)}$ & \texttt{vmo\_14}  & Fraction of trips with average speed exceeding 130 kph\\
        \cmidrule(l){1-3} \\
        $x_{it, 1}^{(m)}$ & \texttt{vma\_1}  & Fraction of trips with maximum speed between 0 and 10 kph\\
        $x_{it, 2}^{(m)}$ & \texttt{vma\_2}  & Fraction of trips with maximum speed between 10 and 20 kph\\
        \vdots & \vdots & \vdots\\
        $x_{it, 16}^{(m)}$ & \texttt{vma\_16}  & Fraction of trips with maximum speed exceeding 150 kph\\
        \cmidrule(l){1-3} \\
        $x_{it, 1}^{(k)}$ & \texttt{d\_1}  & Fraction of trips with distance between 0 and 5 km\\
        $x_{it, 2}^{(k)}$ & \texttt{d\_2}  & Fraction of trips with distance between 5 and 10 km\\
        \vdots & \vdots & \vdots\\
        $x_{it, 10}^{(k)}$ & \texttt{d\_10}  & Fraction of trips with distance exceeding 45 km\\
        \bottomrule
    \end{tabular}
    \caption{Telematics Inputs Names Translation}
    \label{tab:translation}
\end{table}


\end{document}